\documentclass[11pt]{article}
\usepackage[preprint]{acl}
\usepackage{times}
\usepackage{latexsym}
\usepackage[T1]{fontenc}
\usepackage[utf8]{inputenc}
\usepackage{microtype}
\usepackage{inconsolata}
\usepackage{amsmath,amssymb,amsthm}
\usepackage{booktabs}
\usepackage{adjustbox}
\usepackage{graphicx}
\usepackage{xcolor}
\usepackage{algorithm}
\usepackage[noend]{algpseudocode}
\usepackage{tikz}
\usetikzlibrary{arrows.meta,positioning,calc}
\usepackage{pgfplots}
\pgfplotsset{compat=1.17}
\usepackage{enumitem}
\usepackage[most]{tcolorbox}

\definecolor{oiblue}{HTML}{0072B2}
\definecolor{oiorange}{HTML}{E69F00}
\definecolor{oigreen}{HTML}{009E73}
\definecolor{oiverm}{HTML}{D55E00}
\definecolor{oisky}{HTML}{56B4E9}
\definecolor{oipurple}{HTML}{CC79A7}
\usetikzlibrary{shapes.geometric}

\newtheorem{proposition}{Proposition}
\newtheorem{lemma}{Lemma}
\newtheorem{corollary}{Corollary}
\theoremstyle{remark}

\newcommand{\boldhdr}[1]{\vspace{0.05cm}\noindent\textbf{#1}}
\newcommand{\expbold}[1]{\vspace{0.1cm}\noindent\textbf{#1}}

\newcommand{\method}{\textsc{Shift}}
\newcommand{\R}{\mathbb{R}}
\newcommand{\E}{\mathbb{E}}
\newcommand{\Df}{\mathcal{D}_f}
\newcommand{\Dr}{\mathcal{D}_r}
\newcommand{\KL}{\mathrm{KL}}
\newcommand{\thb}{\theta_{\mathrm{b}}}
\newcommand{\tho}{\theta_{0}}
\newcommand{\thr}{\theta_{r}}
\newcommand{\proj}{P_{r}}

\title{Unlearning by Extrapolating a Shrink-and-Heal Lookahead}

\author{Hoang-Chau Luong \ \ \ \ \  Lingwei Chen \\
  Golisano College of Computing and Information Sciences \\
  Rochester Institute of Technology \\
  Rochester, NY, United States \\
  \texttt{cl6300@rit.edu, lwcics@rit.edu}}

\begin{document}
\maketitle

\begin{abstract}
    Unlearning aims to make a large language model behave like one retrained without its forget set. Prevailing likelihood-suppression objectives, however, lack an explicit forgetting target: they continually reward lower forget-set likelihood, making forgetting sensitive to the optimization budget. We address this limitation by estimating the direction toward the oracle without training it. Our key idea is to shrink the fine-tuned model toward its pre-fine-tuning weights and briefly re-learn the retain set, thereby removing the forget-specific contribution; in the linearized regime, this \emph{shrink-and-heal lookahead} moves the model a fixed fraction of the way to the retrained model, its unique fixed point. Our method, \method{}, extrapolates the lookahead's per-token changes into distillation targets and iteratively refines them. Evaluated against eleven baselines under a unified OpenUnlearning protocol, \method{} achieves the highest aggregate score across all TOFU settings and the closest oracle alignment, $3.7\times$ closer on average than the next-best method; on MUSE, it preserves the most retain knowledge. Under larger optimization budgets, baselines overshoot the oracle, while \method{} remains near it, even under relearning attacks.
\end{abstract}

\begin{figure*}[t]
\centering
\begin{minipage}[b]{0.50\textwidth}
\centering
\begin{tikzpicture}
\begin{axis}[width=\linewidth, height=5.1cm, xmode=log, xlabel={learning rate}, ylabel={privacy leakage PL (\%)},
  xtick={5e-6,1e-5,2e-5,3e-5,5e-5}, xticklabels={$5{\cdot}10^{-6}$,$10^{-5}$,$2{\cdot}10^{-5}$,$3{\cdot}10^{-5}$,$5{\cdot}10^{-5}$},
  xmin=4.2e-6, xmax=6e-5, ymin=-108, ymax=68, ytick={-100,-50,0,50},
  legend style={font=\scriptsize, draw=none, fill=none, at={(0.5,1.03)}, anchor=south, legend columns=4,
    /tikz/every even column/.append style={column sep=4pt}},
  tick label style={font=\scriptsize}, label style={font=\small}, grid=major, grid style={gray!15},
  every axis plot/.append style={line width=0.9pt, mark size=1.7pt}]
\fill[oigreen!12] (axis cs:4.2e-6,-25) rectangle (axis cs:6e-5,25);
\draw[oigreen!70!black, densely dashed, line width=0.6pt] (axis cs:4.2e-6,0) -- (axis cs:6e-5,0);
\node[font=\scriptsize\itshape, text=oigreen!50!black, anchor=west] at (axis cs:4.4e-6,14) {near the oracle};
\node[font=\scriptsize\itshape, text=oiverm!80!black, anchor=west, align=left] at (axis cs:4.4e-6,52) {over-forgetting $\uparrow$};
\node[font=\scriptsize\itshape, text=gray!80!black, anchor=east, align=right] at (axis cs:5.6e-5,-62) {under-forgetting $\downarrow$};
\addplot[oiorange, mark=square*] coordinates {(5e-6,-100) (1e-5,-70) (2e-5,8) (3e-5,37) (5e-5,52)};
\addplot[oipurple, mark=triangle*] coordinates {(5e-6,-100) (1e-5,-94) (2e-5,-28) (3e-5,36) (5e-5,54)};
\addplot[oiverm, mark=diamond*] coordinates {(5e-6,-98) (1e-5,-54) (2e-5,49) (3e-5,57) (5e-5,57)};
\addplot[gray, mark=pentagon*] coordinates {(5e-6,-98) (1e-5,-41) (2e-5,50) (3e-5,50) (5e-5,46)};
\addplot[oisky, mark=x] coordinates {(5e-6,-96) (1e-5,-48) (2e-5,14) (3e-5,24) (5e-5,40)};
\addplot[black!45, mark=+] coordinates {(5e-6,-100) (1e-5,-100) (2e-5,-99) (3e-5,-99) (5e-5,-95)};
\addplot[oiblue, line width=2pt, mark=*, mark size=2.4pt] coordinates {(5e-6,-97) (1e-5,-74) (2e-5,-23) (3e-5,6) (5e-5,6)};
\legend{NPO, WGA, GradDiff, RMU, PDU, SimNPO, \textbf{\method{}}}
\end{axis}
\end{tikzpicture}\\[-1pt]
{\small (a) Without a target, forgetting follows the budget.}
\end{minipage}\hfill
\begin{minipage}[b]{0.47\textwidth}
\centering
\begin{tikzpicture}[x=1.1cm, y=1.0cm, every node/.style={font=\scriptsize},
  dot/.style={circle, fill, inner sep=1.5pt}, arr/.style={-{Stealth[length=2mm]}, line width=0.9pt}]
\fill[oiverm!8] (1.9,-1.45) rectangle (5.0,0);
\node[oiverm!80!black, font=\scriptsize\itshape, anchor=south west] at (1.95,-1.43) {forgets more than the oracle};
\draw[->, gray] (-0.25,0) -- (5.0,0);
\node[gray, anchor=north] at (1.0,-0.06) {retain-supported part $\proj\tau$};
\draw[->, gray] (0,-1.45) -- (0,2.95) node[below right, gray, align=left] {forget-specific\\part $(I-\proj)\tau$};
\draw[densely dashed, gray!70] (3,-1.45) -- (3,2.95);
\node[gray, anchor=south west] at (3.05,2.6) {models that fit $\Dr$};
\node[dot, label={[font=\small]left:{$\thb$}}] (b) at (0,0) {};
\node[dot, label={[font=\small]above left:{$\tho$}}] (t) at (3,2.3) {};
\node[dot, oiorange] (s) at (1.5,1.15) {};
\draw[arr, oiorange] (t) -- node[above left=-1pt, oiorange!90!black] {shrink by $\rho$} (s);
\draw[arr, oiorange, densely dashed] (s) -- node[below, oiorange!90!black] {heal on $\Dr$} (2.93,1.15);
\draw[oiblue, line width=1.8pt, -{Stealth[length=2.6mm]}] (3,2.22) -- (3,0.13);
\node[dot, oiblue, inner sep=1.8pt] (p) at (3,1.15) {};
\node[oiorange!90!black, anchor=south east, inner sep=1pt] at (2.95,1.2) {$\Pi(\tho)$};
\node[oiblue, anchor=east, align=right] at (2.9,0.4) {\method{}: extrapolate by $\tfrac{s}{\rho}$,\\look ahead again};
\node[star, star points=5, star point ratio=2.3, fill=oigreen, inner sep=1.5pt] (r) at (3,0) {};
\node[oigreen!50!black, anchor=north west, font=\small] at (3.06,-0.04) {$\thr$ (oracle)};
\draw[arr, oiverm] (3.07,2.32) .. controls (4.6,2.25) and (4.75,0.2) .. (4.4,-0.98);
\node[oiverm!90!black, anchor=east, align=right] at (4.42,-0.7) {likelihood\\suppression};
\end{tikzpicture}\\[-1pt]
{\small (b) The shrink-and-heal lookahead points to the oracle.}
\end{minipage}
\vspace{-0.2cm}
\caption{(a) Privacy leakage on TOFU \texttt{forget05} (Llama-3.2-1B) across learning rates under a fixed 10-epoch budget ($0$: oracle; positive: the forget set looks \emph{less} familiar than never-seen data). Likelihood-suppression methods overshoot the oracle, whereas \method{} stabilizes near it. (b) In the linearized regime (Prop.~\ref{prop:probe}), shrinking $\tho$ toward the base $\thb$ and healing on the retain set removes a fraction $\rho$ of the forget-specific contribution; \method{} extrapolates and iterates this step, which stops at the oracle $\thr$, whereas likelihood suppression has no reason to stop there.}
\label{fig:teaser}
\vspace{-0.2cm}
\end{figure*}
\section{Introduction}

Large language models (LLMs) acquire substantial knowledge through fine-tuning, but may also memorize sensitive or unwanted training data~\citep{carlini2021extracting}. When such data must be removed, retraining from scratch is often prohibitively expensive. This motivates \emph{machine unlearning}, which aims to remove the influence of a \emph{forget set} while preserving what the model learned from the remaining \emph{retain set}~\citep{liu2025rethinking}. Ideally, the unlearned model should behave like an \emph{oracle} trained on the retain set alone. This oracle serves as the reference for evaluating unlearning in established benchmarks~\citep{maini2024tofu,shi2025muse}, and has been advocated as the defining criterion for unlearning~\citep{yoon2026position}.

Most existing unlearning methods are built on gradient ascent over the forget set~\citep{jang2023knowledge,yao2024large}, often combined with retain-set regularization to preserve model utility~\citep{liu2022continual}. Subsequent work bounds or reshapes the ascent objective to prevent model collapse, such as NPO \citep{zhang2024negative}, SimNPO
\citep{fan2024simplicity}, and token reweighting \citep{wang2025rethinking,yang2025exploring}, extends the suppression to the model's own high-confidence alternatives~\citep{li2026llm}, or replaces it with a distillation target
\citep{chundawat2023can,eldan2023whos,dong2025undial,zhong2026duet}.

Despite these refinements, a fundamental yet underexplored limitation remains in likelihood-suppression approaches: \textit{they lack an explicit target for forgetting}. By rewarding lower forget-set likelihood, these objectives do not generally admit the oracle as a stationary point, allowing optimization to proceed beyond the desired forgetting level. Consequently, the extent of forgetting becomes sensitive to the learning rate and number of training steps. As illustrated in Figure~\ref{fig:teaser}(a), varying the optimization budget can shift the same method from insufficient forgetting to over-forgetting, where forget-set examples become less familiar than never-seen data, potentially exposing their former membership~\citep{hayes2025inexact}. This sensitivity is not merely an optimization issue: the objective itself provides no criterion for stopping at the oracle.

An explicit target could address this limitation, and the best target is the oracle itself, which is exactly what one cannot afford to train. Our key observation is that a fine-tuned model already contains a cheap estimate of where the oracle lies. Specifically, shrinking the model toward its pre-fine-tuning weights partially reverses everything learned during fine-tuning, while briefly re-learning the retain set restores knowledge supported by the retained data; what remains removed is the contribution attributable to the forget set. We call this operation the \emph{shrink-and-heal lookahead}. As shown in Figure~\ref{fig:teaser}(b), under linearized fine-tuning dynamics, it moves the model a fixed fraction of the way toward the oracle, which is its unique fixed point.

Building on this observation, we propose \method{} (\textbf{Sh}rink--heal \textbf{I}teration toward the \textbf{F}ixed poin\textbf{T}), an unlearning framework that uses the lookahead to guide prediction-level updates. Rather than adopting the lookahead model directly, \method{} measures its induced changes in forget-set predictions and extrapolates them to estimate an oracle-directed update. The resulting target is then distilled into the model on forget tokens, while its retain-set predictions are anchored to preserve utility. By repeatedly recomputing the lookahead from the updated model, \method{} progressively corrects its approximation to the oracle through a Newton-style iteration.

We evaluate \method{} on TOFU~\citep{maini2024tofu} and MUSE~\citep{shi2025muse} against eleven baselines under a unified protocol. Across the TOFU settings, \method{} achieves the closest forget-set statistics to the oracle, $3.7\times$ closer on average than the next-best method. It also obtains the highest aggregate score among methods that avoid over-forgetting. On MUSE, \method{} preserves the most retain knowledge, while removing more memorization than NPO and SimNPO. Furthermore, \method{} maintains forgetting behavior close to the oracle across different learning rates, demonstrating reduced sensitivity to the optimization budget.

\boldhdr{Contributions.} (1) We identify a fundamental limitation of likelihood-suppression unlearning: without an explicit target, its forgetting behavior depends heavily on the optimization budget (Section~\ref{sec:rethink}). (2) We introduce the shrink-and-heal lookahead and prove that, in the linearized regime, it steps a fixed fraction toward the oracle, which is its unique fixed point, and that distilling its extrapolated per-token change yields a Newton-style update toward the oracle (Section~\ref{sec:method}). (3) We develop \method{} and demonstrate on TOFU and MUSE that it achieves the strongest oracle alignment and the best overall performance among methods that avoid over-forgetting (Section~\ref{sec:exp}).

\section{Related Work}
\label{sec:related}

\textbf{Likelihood suppression.} Early unlearning methods use gradient ascent on the forget set~\citep{jang2023knowledge,yao2024large}, often combined with retain-set regularization~\citep{liu2022continual}. Subsequent approaches introduce bounded or reweighted objectives~\citep{zhang2024negative,fan2024simplicity,
wang2025rethinking,yang2025exploring}, refusal targets~\citep{maini2024tofu}, representation misdirection~\citep{li2024wmdp}, and constrained optimization~\citep{entesari2025constrained}. Recent work also explores adaptive stopping criteria based on the model's own predictions~\citep{digennaro2026fast}. Despite their different mechanisms, they do not explicitly target the behavior of the retain-only oracle, which provides a principled reference for unlearning~\citep{yoon2026position,hayes2025inexact}.

\boldhdr{Targets and distillation.} Distillation-based methods introduce alternative targets to guide forgetting. UNDIAL reduces the gold-token logit by a fixed offset~\citep{dong2025undial}, while ``Who's Harry Potter'' subtracts the excess of a model reinforced on the forget data~\citep{eldan2023whos}. Other approaches construct teachers using
incompetent models~\citep{chundawat2023can}, in-context prompts~\citep{zhong2026duet}, or assistants trained to memorize the forget set~\citep{ji2024reversing}. In contrast, \method{}'s teacher is calibrated by the retain set and has the oracle as its fixed point.

\boldhdr{Checkpoint arithmetic and resets.} The closest related approach is MSA~\citep{rezaei2026revisiting}, which fine-tunes an earlier checkpoint on the forget set and subtracts the resulting parameter difference scaled by a tuned coefficient. \method{} instead estimates the unlearning direction through shrinkage and retain-set recovery, without training on the forget set, and derives its extrapolation scale from the oracle fixed-point property. Related techniques, such as shrink-and-perturb~\citep{ash2020warm} and rewinding~\citep{mu2024rewind} reset model weights, whereas \method{} uses the reset to construct distillation targets. Unlike the Lookahead optimizer~\citep{zhang2019lookahead}, which pulls weights toward a few fast steps, our lookahead is a model-level shrink-and-heal operation rather than an optimizer (see Appendix~\ref{app:related} for further discussion).


\section{Preliminaries and Motivation} \label{sec:prelim}

\textbf{Problem definition and evaluation.} A base model $\thb$ is fine-tuned on $\mathcal{D}=\Df\cup\Dr$, producing the target model $\tho$, where $\Df$ and $\Dr$ denote the forget and retain sets, respectively. Given $\tho$, $\Df$, and $\Dr$, an unlearning method aims to obtain a model $\theta$ that behaves like the oracle $\thr$, defined as $\thb$ fine-tuned on $\Dr$ alone. The oracle is unavailable during unlearning. For a sequence $x$, let $\ell_\theta(\cdot\mid x_{<t})$ denote the next-token log-probabilities at position $t$. Most likelihood-suppression methods optimize an objective of the form:
\begin{equation}
    -\gamma\,\mathcal{L}_f(\theta)+\alpha\,\mathcal{L}_r(\theta)
\end{equation}
where $\mathcal{L}_f$ is the forget-set negative log-likelihood or its bounded transform, $\mathcal{L}_r$ preserves retain-set performance, and \(\gamma,\alpha\) control their contributions.

To evaluate unlearning, on TOFU \citep{maini2024tofu}, we follow OpenUnlearning~\citep{dorna2025openunlearning} and report \emph{Mem.}, the harmonic mean of $1-$extraction strength, $1-$exact memorization, $1-$paraphrased-answer probability, and $1-$truth ratio on $\Df$; \emph{Util.}, the harmonic mean of model utility and fluency; and
\emph{Agg.}, the harmonic mean of Mem. and Util. However, Agg. rewards forgetting monotonically even beyond the oracle, whose own Agg. on \texttt{forget05} is $0.641$. Since unlearning is to match the retrained model~\citep{yoon2026position} and excessive forgetting can itself reveal prior membership~\citep{hayes2025inexact}, we also measure
\emph{distance to the oracle}. Specifically, we compute the absolute difference \(\Delta\) between the unlearned model and the oracle for four forget-set statistics: Mem., truth ratio (TR), extraction strength (ES), and ROUGE-L recall of the answers, and report their
average. We further report \emph{privacy leakage} (PL) that is the relative gap in membership-inference AUC between $\theta$ and $\thr$~\citep{shi2025muse}: PL is $0$ for the oracle, close to $-100$ before unlearning, and positive under over-forgetting. On MUSE \citep{shi2025muse} we report verbatim and knowledge memorization on $\Df$ and knowledge memorization on $\Dr$.

\boldhdr{Motivation: forgetting without a target.} \label{sec:rethink}
To examine how far likelihood suppression forgets, we evaluate six such methods on TOFU \texttt{forget05} using Llama-3.2-1B. Each method is trained for the standard 10 epochs across five learning rates from $5\cdot10^{-6}$ to $5\cdot10^{-5}$. The results are shown in Figure~\ref{fig:teaser}(a), with detailed measurements in Appendix Table~\ref{tab:lr}. Two key observations emerge:
\begin{itemize}[leftmargin=10pt,noitemsep,topsep=4pt]
    \item \textit{Observation 1: forgetting is highly sensitive to the budget.} Five of the six methods reach or surpass the oracle's forgetting level within the tested learning rates. NPO's privacy leakage increases from $-100$ to $+52$, and WGA's from $-100$ to $+54$; GradDiff, RMU, and PDU already over-forget at a learning rate of $2\cdot10^{-5}$, and SimNPO never reaches the oracle. No method lands near the oracle for a reason internal to the method.

    \item \textit{Observation 2: aggregate scores favor over-forgetting.} The strongest baseline scores are achieved beyond the oracle: WGA reaches an Agg. of $0.692$ at PL $=+54$, while NPO peaks at PL \(=+37\). Thus, tuning for Agg. alone pushes a method past the oracle rather than toward it.
\end{itemize}

\noindent The underlying reason lies in the optimization objectives themselves. Their forget-set terms reward decreasing the likelihood of $\Df$ without identifying the oracle as a stopping point; bounded variants such as NPO~\citep{zhang2024negative} attenuate the suppression but do not ensure convergence to the oracle. The extent of forgetting is thus governed by the number of training epochs and the learning rate. An unlearning objective should instead explicitly target an estimate of the oracle.

\begin{figure*}[t]
\centering
\resizebox{\textwidth}{!}{%
\begin{tikzpicture}[font=\scriptsize, >={Stealth[length=1.8mm]},
  panel/.style={draw=gray!45, rounded corners=3pt, fill=gray!3},
  head/.style={font=\footnotesize\bfseries, anchor=north west},
  wbox/.style={draw, rounded corners=2pt, inner sep=2.5pt, align=center, font=\scriptsize},
  arr/.style={->, line width=0.8pt}]
\draw[panel] (0,0) rectangle (3.55,3.55);
\draw[panel] (3.8,0) rectangle (8.45,3.55);
\draw[panel] (8.7,0) rectangle (11.85,3.55);
\draw[panel] (12.1,0) rectangle (15.85,3.55);
\node[circle, fill=black, text=white, font=\tiny\bfseries, inner sep=1.1pt] at (0.25,3.3) {1};
\node[font=\footnotesize\bfseries, anchor=west] at (0.41,3.3) {Look ahead};
\node[circle, fill=black, text=white, font=\tiny\bfseries, inner sep=1.1pt] at (4.05,3.3) {2};
\node[font=\footnotesize\bfseries, anchor=west] at (4.21,3.3) {Read it on forget tokens};
\node[circle, fill=black, text=white, font=\tiny\bfseries, inner sep=1.1pt] at (8.95,3.3) {3};
\node[font=\footnotesize\bfseries, anchor=west] at (9.11,3.3) {Extrapolate};
\node[circle, fill=black, text=white, font=\tiny\bfseries, inner sep=1.1pt] at (12.35,3.3) {4};
\node[font=\footnotesize\bfseries, anchor=west] at (12.51,3.3) {Distill};
\node[wbox, fill=oiblue!10, draw=oiblue] (cur) at (1.78,2.62) {current model $\theta_n$};
\node[wbox, fill=oiorange!10, draw=oiorange] (shr) at (1.78,1.6) {$\theta_n-\rho\,(\theta_n-\thb)$};
\node[wbox, fill=oiorange!22, draw=oiorange] (prb) at (1.78,0.55) {lookahead $\Pi(\theta_n)$};
\draw[arr, oiorange] (cur) -- node[right, font=\tiny, align=left] {shrink toward\\the base} (shr);
\draw[arr, oiorange] (shr) -- node[right, font=\tiny, align=left] {heal $k$ steps\\on $\Dr$} (prb);
\node[anchor=north west, align=left, font=\tiny, text width=4.4cm] at (3.9,3.08)
  {Q: What genre is [author] known for?\\A: \ldots\ known for \textbf{thrillers}};
\begin{scope}[shift={(4.25,0.42)}]
  \foreach \h/\c [count=\i] in {0.70/oiverm, 0.12/oiblue!55, 0.08/oiblue!55}
    \fill[\c] ({0.34*(\i-1)},0) rectangle ({0.34*(\i-1)+0.26},{1.75*\h});
  \draw[gray] (-0.08,0) -- (1.0,0);
  \node[below, font=\scriptsize] at (0.47,0) {$p_{\theta_n}$};
\end{scope}
\begin{scope}[shift={(6.55,0.42)}]
  \foreach \h/\c [count=\i] in {0.40/oiverm, 0.25/oiblue!55, 0.18/oiblue!55}
    \fill[\c] ({0.34*(\i-1)},0) rectangle ({0.34*(\i-1)+0.26},{1.75*\h});
  \draw[gray] (-0.08,0) -- (1.0,0);
  \node[below, font=\scriptsize] at (0.47,0) {$p_{\Pi}$};
\end{scope}
\draw[arr, gray] (5.45,1.0) -- node[above, font=\tiny] {$\ell_\Pi-\ell_{n}$} (6.35,1.0);
\fill[oiverm] (3.98,2.3) rectangle (4.1,2.42); \node[anchor=west, font=\tiny] at (4.1,2.36) {gold token};
\fill[oiblue!55] (5.3,2.3) rectangle (5.42,2.42); \node[anchor=west, font=\tiny] at (5.42,2.36) {alternatives};
\node[anchor=north, align=center] at (10.27,3.08)
  {$\ell_T=\ell_n+\tfrac{s}{\rho}\,(\ell_\Pi-\ell_n)$\\[-1pt]{\tiny renormalized on top tokens}};
\begin{scope}[shift={(9.8,0.42)}]
  \foreach \h/\c [count=\i] in {0.18/oiverm, 0.42/oiblue!55, 0.32/oiblue!55}
    \fill[\c] ({0.34*(\i-1)},0) rectangle ({0.34*(\i-1)+0.26},{1.75*\h});
  \draw[gray] (-0.08,0) -- (1.0,0);
  \node[below, font=\scriptsize] at (0.47,0) {teacher $p_T$};
\end{scope}
\node[align=center, font=\tiny, text=gray!60!black] at (10.27,2.05) {how much to forget \emph{and}\\what to say instead};
\node[wbox, fill=oigreen!10, draw=oigreen!70!black, text width=3.3cm] (lf) at (13.97,2.55)
  {forget tokens: $\KL\big(p_T\,\Vert\,p_\theta\big)$};
\node[wbox, fill=gray!10, draw=gray, text width=3.3cm] (lr) at (13.97,1.6)
  {retain anchor: $\alpha\,\KL\big(p_{\tho}\,\Vert\,p_\theta\big)$};
\node[wbox, fill=oiblue!10, draw=oiblue] (nxt) at (13.97,0.55) {$\theta_{n+1}$};
\draw[arr] (lr) -- (nxt);
\draw[arr, oiorange] (prb.east) -- (4.12,0.55);
\draw[arr] (7.6,1.0) -- (9.68,1.0);
\draw[arr] (10.9,1.0) -- (11.5,1.0) |- (lf.west);
\draw[arr, oiblue] (nxt.south) -- ++(0,-0.55) -- node[below, font=\scriptsize] {next round: look ahead from $\theta_{n+1}$ (the lookahead is discarded)} (-0.25,-0.33) |- (cur.west);
\end{tikzpicture}%
}
\caption{One round of \method{}. (1) Shrink toward the pre-fine-tuning weights and re-fit the retain set. (2) On forget
answers, the lookahead lowers the memorized token and raises the alternatives the retain set supports. (3) Extrapolating by
$s/\rho$ gives the teacher, the oracle's next-token distribution in the linearized regime. (4) Distill it on forget
tokens, anchored to $\tho$ on retain data; the next round looks ahead from the new model.}
\label{fig:loop}
\end{figure*}
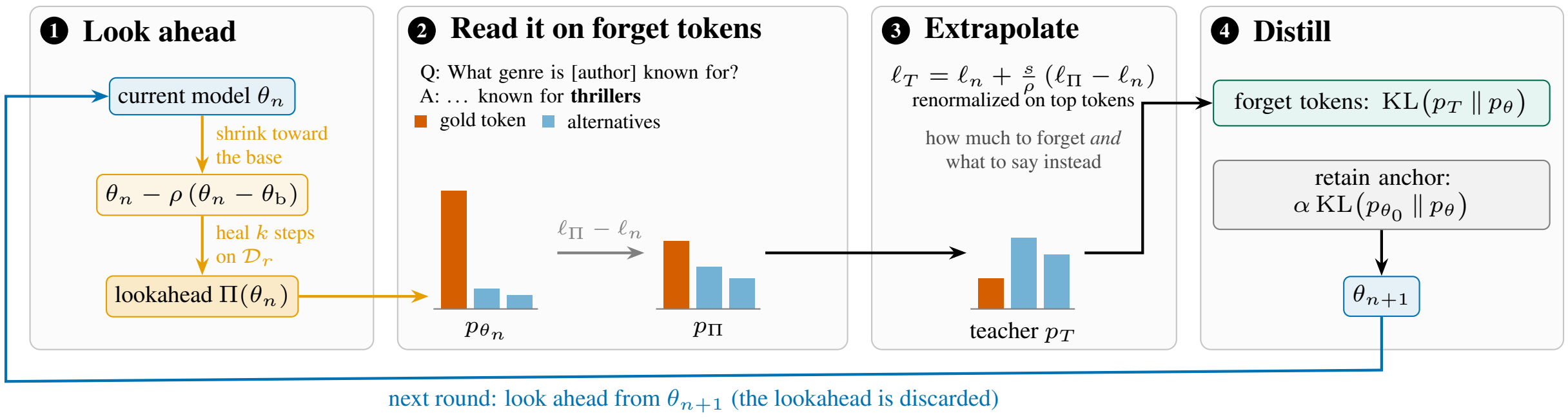


\section{\method{}: Unlearning toward an Estimated Oracle}
\label{sec:method}

The oracle provides a target for unlearning, but training it would defeat the purpose of unlearning. Instead, we use the fine-tuned model $\tho$, the base model $\thb$, and the retain set $\Dr$ to estimate the direction toward the oracle at low computational cost. We first introduce the \emph{shrink-and-heal lookahead} and analyze its fixed point (Section~\ref{sec:probe}). We then extrapolate its prediction changes to construct distillation targets on forget tokens (Section~\ref{sec:teacher}) and iteratively refine these targets as the model evolves (Section~\ref{sec:iterate}). Figure~\ref{fig:loop} illustrates one round.

\subsection{Shrink-and-Heal Lookahead}
\label{sec:probe}

Fine-tuning displaces the base model by $\tau=\tho-\thb$ that encodes both $\Dr$ and
$\Df$. Pulling the model back toward $\thb$ by a fraction $\rho$ removes a $\rho$-share of everything that fine-tuning added. Re-fitting the retain set for a few steps then restores the component supported by $\Dr$. What remains removed is the part of $\tau$ that $\Dr$ cannot explain, i.e., what only $\Df$ taught. We formalize this operation as the \emph{shrink-and-heal lookahead}:
\begin{equation}
\Pi(\theta)=\mathrm{Heal}_k\big(\theta-\rho\,(\theta-\thb)\big),\label{eq:probe-def}
\end{equation}
where $\mathrm{Heal}_k$ denotes $k$ optimization steps on $\Dr$.

To characterize this operation, we consider the linearized fine-tuning regime~\citep{jacot2018neural,malladi2023kernel,ortizjimenez2023task}, which has recently been extended to attention networks initialized from pre-trained weights~\citep{sun2026theoretical}. In this model, the logits satisfy $f(x;\theta)=f(x;\thb)+J_x(\theta-\thb)$, and gradient descent from $\thb$ reaches the minimum-norm displacement that fits its data~\citep{gunasekar2018characterizing}. Let $J_r$ stack the retain-set Jacobians and let $\proj=J_r^{+}J_r$ project onto their row space. $\proj\tau$ and $(I-\proj)\tau$ represent the retain-supported and forget-specific components of the fine-tuning displacement, respectively.

\begin{proposition}[The lookahead is a $\rho$-step toward the oracle]\label{prop:probe}
If the heal runs to convergence, then
\begin{align}
\thr &= \thb + \proj\,\tau, \label{eq:oracle}\\
\Pi(\tho) &= \tho+\rho\,(\thr-\tho). \label{eq:probe}
\end{align}
More generally, every $\theta=\thb+\tau'$ that fits $\Dr$ satisfies $\Pi(\theta)-\theta=-\rho\,(I-\proj)\,\tau'$, so the oracle $\thr$ is the unique fixed point of $\Pi$.
\end{proposition}

\noindent The heal moves only within the row space of $J_r$, so it restores exactly the retain-supported component $\rho\proj\tau$ of what the shrink removed; the net move $-\rho(I-\proj)\tau$ is a fraction $\rho$ of the forget-specific component of fine-tuning (Figure~\ref{fig:teaser}(b); proof in Appendix~\ref{app:proofs}). The lookahead therefore points to the oracle without training it, and needs no knowledge of which parameters encode $\Df$. At the oracle, it produces no further change, which is the property that likelihood suppression lacks (Section~\ref{sec:rethink}).

\boldhdr{What the lookahead does in practice.} Real fine-tuning is not linear and the heal is short. We therefore examine the lookahead on TOFU \texttt{forget10} with Llama-3.2-1B ($\rho=\tfrac12$, $k=32$; Appendix~\ref{app:stage0}) before building on it, comparing gold-token log-probabilities before and after the lookahead, extrapolated by $1/\rho$. \textit{(1) The lookahead isolates the forget set}: its forget-specific drop is $1.9$ nats per token, approximately $80\%$ of the true gap between $\tho$ and the oracle ($2.45$ nats, computed post hoc). \textit{(2) It also moves unrelated text}: on never-seen authors, it raises the gold tokens by $0.96$ nats, as shrinking makes $\tho$ less over-confident in general, and this drift does not shrink as $k$ increases from $16$ to $64$.

\boldhdr{Design requirements.} These observations give three requirements. First, the lookahead should be \emph{read, not kept}: it covers only a fraction $\rho$ of the way, and changes predictions everywhere, including where unlearning should not act. Second, its signal should act \emph{only where unlearning applies}, the forget tokens, while the rest of the model's behavior is held fixed. Third, the estimate should be \emph{refreshed}: since the lookahead is accurate only locally, it should be recomputed from the updated model. The next two subsections instantiate them.

\subsection{From Lookahead to Teacher}
\label{sec:teacher}

\textbf{Reading the lookahead in output space.} Following the first requirement, we never use $\Pi(\theta)$ as a model, but instead only compare its predictions with those of the current model. Proposition~\ref{prop:probe} shows how to turn this comparison into the oracle's predictions. Extrapolating the lookahead step by $1/\rho$ lands on the oracle: $\tho+\tfrac{1}{\rho}\big(\Pi(\tho)-\tho\big)=\thr$. Since logits are linear in $\theta$, we can obtain: 
\begin{equation}
f(x;\thr)=f(x;\tho)+\tfrac{1}{\rho}\Big(f\big(x;\Pi(\tho)\big)-f(x;\tho)\Big).
\label{eq:extrap-logits}
\end{equation}
Therefore, two forward passes suffice to recover the oracle's logits, without the oracle. In practice, we store log-probabilities rather than logits, as the softmax is invariant to the per-position normalization (see Lemma~\ref{lem:norm} in Appendix~\ref{app:proofs}).

\boldhdr{Extrapolated teacher.} At each answer position $t$ of a forget example, let $\ell_{n,t}$ and $\ell_{\Pi,t}$ be the log-probabilities of the current model $\theta_n$ and its lookahead, restricted to $V_t$, the union of both models' top-256 tokens. We form the teacher by:
\begin{equation}
\log p_{T,t}=\mathrm{norm}_{V_t}\!\Big(\ell_{n,t}+\mathrm{clip}_{\kappa}\big(\tfrac{s_n}{\rho}(\ell_{\Pi,t}-\ell_{n,t})\big)\Big),
\label{eq:teacher}
\end{equation}
which renormalizes the probabilities within $V_t$ and leaves the current model's mass outside $V_t$ unchanged. The step $s_n$ scales the extrapolation: $s_n=\rho$ reproduces the lookahead, and $s_n=1$ takes the full step in Eq.~(\ref{eq:extrap-logits}). By Lemma~\ref{lem:norm}, the teacher at $s_n=1$ is the oracle's next-token distribution on $V_t$ in the linearized regime. We clip each extrapolated change at $\kappa=10$ nats, since dividing by $\rho$ amplifies noise in the lookahead.

The change $\ell_{\Pi,t}-\ell_{n,t}$ does not only lower the memorized token; it also raises the alternatives that a model without $\Df$ would predict (Figure~\ref{fig:loop}, panel 2). This distinction matters because suppressing only the gold token can transfer its probability mass to semantically related continuations~\citep{li2026llm}. Unlike methods that apply a fixed reduction to the gold-token logit~\citep{dong2025undial}, our teacher tells the model both \emph{how much} to forget at each position and \emph{what to say instead}, as estimated from the retain set.

\subsection{Distillation and Iteration}
\label{sec:iterate}

\paragraph{Distillation with a retain anchor.} Following the second requirement, the teacher acts only on forget tokens, and every retain example anchors the model to $\tho$'s own predictive distribution:
\begin{equation}
\mathcal{L}(\theta)=\gamma\,\E_{\Df}\KL\big(p_T\Vert p_\theta\big)+\alpha\,\E_{\Dr}\KL\big(p_{\tho}\Vert p_\theta\big).
\label{eq:loss}
\end{equation}
Unlike the likelihood-suppression objectives, this loss is minimized at the teacher, so additional optimization brings the model closer to the estimated oracle rather than past it. The KL anchor preserves the full distribution on retain data, not only its gold tokens~\citep{zhang2024negative,dong2025undial}, and also holds the retain knowledge that a finite heal fails to restore (Appendix~\ref{app:proofs}). We cache $\tho$'s top-64 tokens per retain position once, avoiding the need to keep a separate reference model in memory. Although the loss acts on outputs, in the linearized regime it carries out exactly the parameter-space step that the lookahead defines:

\begin{proposition}[Distillation implements the step]\label{prop:distill}
Let $\theta_n$ fit $\Dr$, and let the teacher be Eq.~(\ref{eq:teacher}) at step $s$ without clipping or truncation. Gradient descent on the logit-matching form of Eq.~(\ref{eq:loss}) from $\theta_n$ converges to
$\theta_n+\tfrac{s}{\rho}\big(\Pi(\theta_n)-\theta_n\big)$; at $s=1$, this point is the oracle $\thr$.
\end{proposition}

\noindent The proof is provided in Appendix~\ref{app:proofs}. Both loss terms target the prediction outputs of the extrapolated parameter point, and gradient descent updates the parameters only within the row space of the training Jacobian, which contains the desired step.

\boldhdr{Iterating the update.} Following the third requirement, each round takes a new lookahead from the model it has just produced. In the linearized setting, this iteration provably converges to the oracle:

\begin{corollary}[Iteration]\label{cor:iter}
Let $\theta_{n+1}=\theta_n+\tfrac{s}{\rho}\big(\Pi(\theta_n)-\theta_n\big)$ with $\theta_0=\tho$. Every iterate fits $\Dr$, its retain-supported part $\proj(\theta_n-\thb)$ stays equal to $\proj\tau$, and its forget-specific part shrinks geometrically:
\begin{equation}
(I-\proj)(\theta_{n}-\thb)=(1-s)^{n}\,(I-\proj)\,\tau .
\end{equation}
Hence $\theta_n\to\thr$ for every $s\in(0,2)$, and $s=1$ reaches $\thr$ in one step.
\end{corollary}

\noindent The proof is provided in Appendix~\ref{app:proofs}. The step $s=1$ is thus a Newton step on the fixed-point equation $\Pi(\theta)=\theta$, and by Proposition~\ref{prop:distill} each round of distillation realizes one such step. On real models, each new lookahead corrects what the previous linear estimate missed. We use a fixed schedule $s=(\rho,1,1,1)$: the first round takes the lookahead itself without extrapolation, as the lookahead moves $\tho$ the most and extrapolating that first change is the least reliable step (the mean teacher drop at $\tho$ is $0.62$ nats at $s=\tfrac12$ but $2.47$ at $s=1$). The complete procedure is summarized in Algorithm~\ref{alg:shift}.

\boldhdr{What the idealization leaves out.} Each gap between the analysis and practice is covered by one component. A
finite heal restores retain knowledge only partially (Appendix~\ref{app:proofs}), and the lookahead drifts on unrelated text:
the teacher therefore acts only on forget tokens, and the retain anchor holds the rest. Real fine-tuning is nonlinear,
so the lookahead is accurate only locally: the iteration re-linearizes every round.

\section{Experiments}
\label{sec:exp}

\subsection{Setup}

\textbf{Benchmarks and models.} We use TOFU \citep{maini2024tofu} with Llama-3.2-1B-Instruct, Llama-3.2-3B-Instruct and Llama-3.1-8B-Instruct on the \texttt{forget01}, \texttt{forget05} and \texttt{forget10} splits, with the OpenUnlearning target and oracle models~\citep{dorna2025openunlearning}; and MUSE-News \citep{shi2025muse} with Llama-2-7B. The bases $\thb$ are the corresponding instruction-tuned (TOFU) or pre-trained (MUSE) checkpoints.

\expbold{Baselines.} We compare \method{} against eleven unlearning baselines: GradAscent, GradDiff, NPO, SimNPO, DPO with refusal targets (IdkDPO)~\citep{rafailov2023direct}, RMU~\citep{li2024wmdp}, UNDIAL~\citep{dong2025undial}, WGA~\citep{wang2025rethinking}, SatImp~\citep{yang2025exploring}, CE-U~\citep{yang2025ceu} and PDU~\citep{entesari2025constrained}, with the OpenUnlearning implementations.

\expbold{Protocol.} Every baseline follows the OpenUnlearning training configuration \citep{dorna2025openunlearning}
with its default hyperparameters: learning rate $10^{-5}$, 10 epochs, effective batch 32 on TOFU, bf16 weights, one seed,
and the final model (no checkpoint selection). \method{} uses the same learning rate, batch and model selection, with
$\rho=\tfrac12$, $k=16$, $\kappa=10$, $\alpha=\gamma=1$ and four rounds of four epochs. This budget is chosen once, on
Llama-3.2-1B \texttt{forget05}, and used unchanged for every model, split and benchmark in the main results. Appendix~\ref{app:impl} gives the implementation details.

\begin{table}[t]
\centering
\footnotesize
\setlength{\tabcolsep}{3pt}
\setlength{\tabcolsep}{2.6pt}
\begin{adjustbox}{max width=\columnwidth}\begin{tabular}{lrrrrrr}
\toprule
Method & $\Delta$Mem.$\downarrow$ & $\Delta$TR$\downarrow$ & $\Delta$ES$\downarrow$ & $\Delta$RL$\downarrow$ & Avg.$\downarrow$ & FQ$\uparrow$ \\
\midrule
GradAscent & 0.248 & 0.276 & 0.066 & 0.218 & 0.202 & $-89.16$ \\
GradDiff & 0.208 & 0.195 & 0.085 & 0.131 & 0.155 & $-47.38$ \\
NPO & \underline{0.091} & \underline{0.045} & 0.059 & 0.086 & \underline{0.071} & \underline{$-3.29$} \\
SimNPO & 0.414 & 0.107 & 0.470 & 0.339 & 0.332 & $-11.49$ \\
IdkDPO & 0.250 & 0.071 & 0.262 & 0.275 & 0.215 & $-6.67$ \\
RMU & 0.158 & 0.048 & 0.088 & 0.149 & 0.111 & $-3.30$ \\
UNDIAL & 0.131 & 0.090 & 0.091 & \underline{0.064} & 0.094 & $-12.39$ \\
WGA & 0.202 & 0.089 & 0.127 & 0.148 & 0.141 & $-11.37$ \\
SatImp & 0.491 & 0.110 & 0.713 & 0.450 & 0.441 & $-12.26$ \\
CE-U & 0.217 & 0.126 & 0.048 & 0.302 & 0.173 & $-41.69$ \\
PDU & 0.136 & 0.061 & \underline{0.039} & 0.190 & 0.106 & $-12.30$ \\
\midrule
\textbf{SHIFT (ours)} & \textbf{0.019} & \textbf{0.021} & \textbf{0.006} & \textbf{0.029} & \textbf{0.019} & $\mathbf{-2.07}$ \\
\bottomrule
\end{tabular}
\end{adjustbox}
\caption{Distance to the oracle, $|x-x_{\textit{Retain}}|$, for Mem., truth ratio (TR), extraction strength (ES) and
ROUGE-L (RL) on the forget set, and TOFU's forget quality (FQ: $\log_{10}p$ of a test against the oracle's truth
ratios; $0$ is the oracle), averaged over the settings of the Avg.\ column of Table~\ref{tab:agg}. \textbf{Bold} /
\underline{underlined}: best / second best.}
\label{tab:fidelity}
\end{table}

\begin{table*}[t]
\centering
\footnotesize
\setlength{\tabcolsep}{2.2pt}
\begin{adjustbox}{max width=\textwidth}\begin{tabular}{lrrrrrrrrrr}
\toprule
 & \multicolumn{3}{c}{Llama-3.2-1B} & \multicolumn{3}{c}{Llama-3.2-3B} & \multicolumn{3}{c}{Llama-3.1-8B} & \\
\cmidrule(lr){2-4} \cmidrule(lr){5-7} \cmidrule(lr){8-10}
Method & \texttt{forget01} & \texttt{forget05} & \texttt{forget10} & \texttt{forget01} & \texttt{forget05} & \texttt{forget10} & \texttt{forget01} & \texttt{forget05} & \texttt{forget10} & Avg. \\
\midrule
\textit{Full} & 0.118 & 0.154 & 0.159 & 0.024 & 0.059 & 0.063 & 0.005 & 0.016 & 0.011 & 0.068 \\
\textit{Retain} & 0.613 & 0.641 & 0.643 & 0.632 & 0.648 & 0.650 & 0.619 & 0.641 & 0.649 & 0.637 \\
\midrule
GradAscent & 0.458 & 0.214 & 0.000 & 0.409 & \underline{0.567} & 0.000 & 0.617 & 0.000 & 0.000 & 0.252 \\
GradDiff & 0.440 & 0.527 & 0.521 & 0.378 & 0.487 & 0.482 & 0.565 & 0.111 & 0.423 & 0.437 \\
NPO & 0.458 & 0.548 & \underline{0.568} & 0.419 & 0.561 & 0.584 & 0.599 & \textbf{0.666} & \underline{0.650} & \underline{0.561} \\
SimNPO & 0.218 & 0.229 & 0.235 & 0.151 & 0.165 & 0.181 & 0.250 & 0.366 & 0.383 & 0.242 \\
IdkDPO & 0.294 & 0.442 & 0.488 & 0.196 & 0.397 & 0.431 & 0.478 & 0.549 & 0.531 & 0.423 \\
RMU & 0.509 & \underline{0.575} & 0.221 & 0.212 & 0.539 & \textbf{0.646} & 0.533 & 0.612 & 0.447 & 0.477 \\
UNDIAL & 0.456 & 0.480 & 0.484 & 0.446 & 0.532 & 0.546 & \underline{0.630} & 0.649 & \textbf{0.653} & 0.542 \\
WGA & 0.434 & 0.488 & 0.506 & 0.405 & 0.453 & 0.485 & 0.552 & 0.589 & 0.436 & 0.483 \\
SatImp & 0.139 & 0.175 & 0.182 & 0.036 & 0.080 & 0.082 & 0.106 & 0.143 & 0.207 & 0.128 \\
CE-U & 0.557 & 0.000 & 0.000 & 0.538 & 0.000 & 0.000 & 0.282 & 0.000 & 0.000 & 0.153 \\
PDU & \underline{0.573} & 0.335 & 0.499 & \underline{0.559} & 0.045 & 0.520 & 0.018 & 0.631 & 0.310 & 0.388 \\
\midrule
\textbf{SHIFT (ours)} & \textbf{0.619} & \textbf{0.623} & \textbf{0.624} & \textbf{0.635} & \textbf{0.639} & \underline{0.634} & \textbf{0.639} & \underline{0.650} & 0.631 & \textbf{0.633} \\
\bottomrule
\end{tabular}
\end{adjustbox}
\caption{Agg.$\uparrow$ on TOFU, every method at the OpenUnlearning learning rate $10^{-5}$ (one seed). Agg.\ is the
harmonic mean of Mem.\ (forgetting of $\Df$) and Util.\ (retained utility), so it is high only when a method both
forgets and keeps utility.
\textbf{Bold} / \underline{underlined}: best / second-best Agg.\ of each setting.
\textit{Full}: before unlearning; \textit{Retain}: the oracle. Mem.\ and Util.\ are in Appendix
Table~\ref{tab:overview}.}
\label{tab:agg}
\end{table*}

\subsection{Main Results}

\expbold{\method{} ends closest to the oracle.} The goal of unlearning is the retrained model, so we first ask how close each
method ends to it. The results are reported in Table~\ref{tab:fidelity}. \method{} is the closest in every setting but Llama-3.1-8B \texttt{forget01} and, on average, on each of the four statistics: within $0.019$ of the oracle, $3.7\times$ closer than the next method (NPO), with memorization and utility on average within $0.019$ and $0.016$ of the oracle's (Appendix Table~\ref{tab:overview}).
TOFU's own test against the oracle agrees: \method{} has the highest forget quality on average ($-2.07$, against
$-3.29$ for NPO and $-3.30$ for RMU).
The baselines miss the oracle in one of two ways. Most stop short of it: NPO, UNDIAL, WGA, IdkDPO, SimNPO and
SatImp leave Mem.\ $0.09$--$0.49$ below the oracle's. The others lose utility: Util.\
falls $0.16$--$0.56$ below the oracle's for RMU, GradDiff, PDU, GradAscent and CE-U.

\expbold{\method{} has the best aggregate score.} \method{} has the highest Agg.\ in six of the nine settings
(Table~\ref{tab:agg}), $0.009$--$0.076$ above the strongest baseline, and the best average ($0.633$ vs.\ $0.561$ for
NPO). Two of the exceptions go to methods that over-forget there: RMU on Llama-3.2-3B \texttt{forget10} (privacy
leakage $+30$) and NPO on Llama-3.1-8B \texttt{forget05} ($+44$). On Llama-3.1-8B \texttt{forget10}, UNDIAL leads
($0.653$ vs.\ $0.631$), while \method{} stays three times closer to the oracle ($0.012$ vs.\ $0.036$).
\method{} earns its scores by matching the oracle: its memorization stays within $0.045$ of the oracle's in every
setting, even where its Agg.\ exceeds the oracle's own. No baseline is competitive everywhere: the strongest one changes
across settings (PDU, RMU, NPO, GradAscent, UNDIAL), and one to four fall below $0.2$ in each. Our
baseline scores match those reported for tuned baselines (NPO: $0.55$ and $0.57$ on Llama-3.2-1B
\texttt{forget05} and \texttt{forget10}, against $0.54$ and $0.58$ in \citealt{li2026llm}).

\begin{figure}[t]
\centering
\input{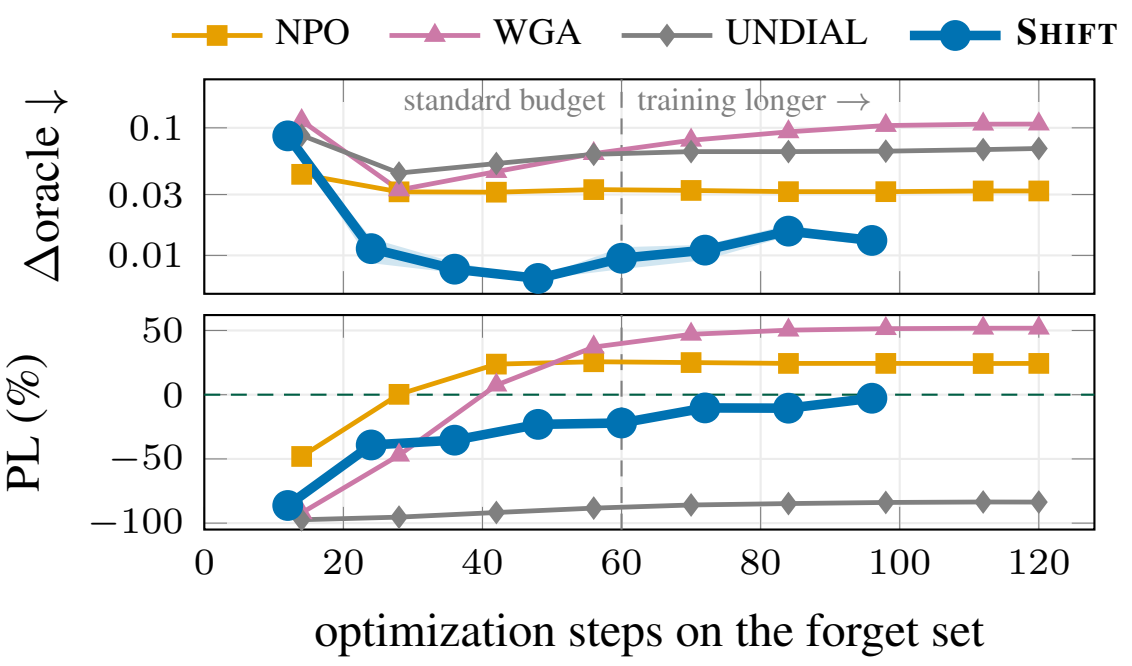}
\begin{tikzpicture}
\pgfplotsset{every axis/.append style={width=0.97\columnwidth, height=3.0cm, xmin=0, xmax=128,
    tick label style={font=\scriptsize}, label style={font=\small}, grid=major, grid style={gray!15},
    every axis plot/.append style={line width=0.9pt, mark size=1.5pt}},
  trajNPO/.style={oiorange, mark=square*}, trajWGA/.style={oipurple, mark=triangle*},
  trajUNDIAL/.style={gray, mark=diamond*}, trajSHIFT/.style={oiblue, line width=1.8pt, mark=*, mark size=2pt}}
\begin{axis}[name=top, ylabel={$\Delta$oracle $\downarrow$}, ymode=log, ymin=0.005, ymax=0.24,
  ytick={0.01,0.03,0.1}, yticklabels={$0.01$,$0.03$,$0.1$}, xticklabels={},
  legend style={font=\scriptsize, draw=none, fill=none, at={(0.5,1.02)}, anchor=south, legend columns=4,
    /tikz/every even column/.append style={column sep=5pt}}]
\draw[densely dashed, gray] (axis cs:60,0.005) -- (axis cs:60,0.24);
\node[font=\tiny, text=gray, anchor=north east, inner sep=1.5pt] at (axis cs:59,0.24) {standard budget};
\node[font=\tiny, text=gray, anchor=north west, inner sep=1.5pt] at (axis cs:61,0.24) {training longer $\rightarrow$};
\trajgap
\legend{NPO, WGA, UNDIAL, \textbf{\method{}}}
\end{axis}
\begin{axis}[at={(top.south)}, anchor=north, yshift=-4pt, xlabel={optimization steps on the forget set},
  ylabel={PL (\%)}, ymin=-105, ymax=62, ytick={-100,-50,0,50}]
\draw[oigreen!60!black, densely dashed] (axis cs:0,0) -- (axis cs:128,0);
\draw[densely dashed, gray] (axis cs:60,-105) -- (axis cs:60,62);
\trajpl
\end{axis}
\end{tikzpicture}
\caption{What happens when training continues, on TOFU \texttt{forget05} (Llama-3.2-1B, learning rate
$2\cdot10^{-5}$): the baselines train for 20 epochs (warm-up, then linear decay over the 20 epochs) and \method{} for
8 rounds. The dashed line marks the end of a standard 10-epoch run (step 60); \method{}'s standard 4 rounds end at step 48.
Top: distance to the oracle (log scale); bottom: privacy leakage (0 is the oracle, positive is over-forgetting).
\method{}: mean over three seeds, shaded from minimum to maximum; baselines: one run, evaluated every second epoch.}
\label{fig:traj}
\end{figure}

\begin{table}[t]
\centering
\footnotesize
\setlength{\tabcolsep}{3pt}
\begin{adjustbox}{max width=\columnwidth}\begin{tabular}{lrrr}
\toprule
Method & VerbMem & KnowMem-F & KnowMem-R \\
\midrule
\textit{Target} & 0.579 & 0.644 & 0.555 \\
\textit{Retrain} & 0.202 & 0.328 & 0.560 \\
\midrule
GradAscent & 0.000 & 0.000 & 0.000 \\
GradDiff & 0.022 & 0.340 & 0.121 \\
NPO & 0.441 & 0.627 & 0.496 \\
SimNPO & 0.519 & 0.621 & 0.496 \\
RMU & 0.276 & 0.461 & 0.476 \\
UNDIAL & 0.138 & 0.130 & 0.204 \\
WGA & 0.079 & 0.245 & 0.228 \\
\midrule
\textbf{SHIFT (ours)} & 0.235 & 0.572 & 0.539 \\
\bottomrule
\end{tabular}
\end{adjustbox}
\caption{MUSE-News, Llama-2-7B, learning rate $10^{-5}$ (baselines: 10 epochs; \method{}: 4 rounds of 4 epochs).
\textit{Target}: before unlearning; \textit{Retrain}: retrained without the forget set (the oracle).}
\label{tab:muse}
\end{table}

\expbold{MUSE.} On MUSE-News (results are reported in Table~\ref{tab:muse}), where the forget set consists of raw text and the base is a pre-trained model, the methods that push verbatim memorization below the retrained oracle (GradAscent, GradDiff, UNDIAL, WGA) substantially degrade retain knowledge (KnowMem-R at most $0.23$). In contrast, \method{} achieves the highest retain-knowledge score ($0.539$, $97\%$ of the target's), while removing more memorization than NPO and SimNPO. It also comes closest to the retrained oracle in verbatim memorization and retain knowledge.

\subsection{Analysis}
\label{sec:analysis}

\expbold{More training does not overshoot.} This analysis and the relearning attacks below use learning rate
$2\cdot10^{-5}$, where the baselines' 10 epochs carry twice the default update (\method{}: $k=32$, two epochs a
round). Figure~\ref{fig:traj} trains NPO, WGA, UNDIAL and \method{} twice as long, the baselines for
20 epochs under their own warm-up and decay, and evaluates them along the way. NPO and WGA cross the oracle within a
standard run: NPO settles at a privacy leakage of $+24$, and WGA keeps rising to $+52$, three times further from the oracle
than at its closest.
UNDIAL never approaches the oracle (privacy leakage at most $-84$). \method{} approaches the oracle and stays there:
its distance is smallest at round 4 ($0.007$) and remains below $0.02$ through round 8, and its privacy leakage rises
toward $0$ without crossing it, as expected of a loss minimized at its teacher rather than at an ever lower
likelihood. Across learning rates (Figure~\ref{fig:teaser}(a), Appendix Table~\ref{tab:lr}), every baseline but SimNPO
overshoots to $+40$--$+57$ at $5\cdot10^{-5}$, whereas \method{} stays within $[-23,+6]$ from $2\cdot10^{-5}$ up.

\expbold{Training cost.} \method{} costs only modestly more than the baselines. Each round adds $k=16$ retain
steps and two forward passes over $\Df$, and distillation runs no reference model, since the teacher and the retain
anchor are cached (Algorithm~\ref{alg:shift}). On one GPU, \method{} trains on Llama-3.2-1B \texttt{forget05} in
19 minutes against NPO's 15.

\begin{table}[t]
\centering
\footnotesize
\setlength{\tabcolsep}{3pt}
\begin{adjustbox}{max width=\columnwidth}\begin{tabular}{lrrrr}
\toprule
Variant & Agg.$\uparrow$ & Mem.$\uparrow$ & Util.$\uparrow$ & $\Delta$oracle$\downarrow$ \\
\midrule
\textbf{SHIFT} & 0.623 & 0.558 & 0.705 & 0.016 \\
\midrule
\quad no heal ($k{=}0$) & 0.617 & 0.553 & 0.698 & 0.015 \\
\quad no extrapolation ($s{=}\rho$) & 0.609 & 0.537 & 0.702 & 0.023 \\
\quad no raised tokens & 0.563 & 0.470 & 0.703 & 0.060 \\
\midrule
Lookahead as the model, 1 round & 0.520 & 0.425 & 0.668 & 0.084 \\
Lookahead as the model, 4 rounds & 0.559 & 0.545 & 0.573 & 0.013 \\
\bottomrule
\end{tabular}
\end{adjustbox}
\caption{Ablation studies on TOFU \texttt{forget05}, Llama-3.2-1B (setting of Table~\ref{tab:agg}; $\Delta$oracle as in
Table~\ref{tab:fidelity}). Raised tokens drive the forgetting, extrapolation saves budget, the heal anchors the target
(Table~\ref{tab:ablation2e5}), and distillation keeps utility: only together are they best in Agg., Mem.\ and Util.}
\label{tab:ablation}
\end{table}

\expbold{Ablations.} Table~\ref{tab:ablation} removes one component at a time, in the setting of
Table~\ref{tab:agg}; every variant falls below \method{} in Agg., Mem.\ and Util.\ alike. \emph{The heal keeps the target at the oracle}:
without it ($k=0$) the lookahead is a pure shrink toward $\thb$ rather than a step toward the oracle
(Proposition~\ref{prop:probe}). At this budget the two end close (Agg.\ $0.617$; Appendix Table~\ref{tab:heal} varies
$k$), but at twice the learning rate, removing the heal nearly doubles the distance to the oracle (Appendix
Table~\ref{tab:ablation2e5}). \emph{Extrapolation saves budget}: the
lookahead without extrapolation ($s=\rho$) moves in the right direction but more slowly (Corollary~\ref{cor:iter}),
reaching a lower Agg.\ ($0.609$) and a larger distance to the oracle ($0.023$) in the same rounds. \emph{Both directions
of the teacher matter}: keeping only the lowered tokens cuts Agg.\ to $0.563$ and leaves the model nearly $4\times$ further
from the oracle, since the raised alternatives are what moves the mass \citep{wu2026unlearning}. \emph{Reading beats
committing}: using the lookahead itself as the model, as resetting methods do \citep{ash2020warm,mu2024rewind}, either stops
short (one round: Agg.\ $0.520$) or, iterated, loses utility ($0.573$ vs.\ $0.705$).

The results of additional analyses, including learning-rate sensitivity (Table~\ref{tab:lr}), heal-length sensitivity (Table~\ref{tab:heal}), and ablations at a higher learning rate (Table~\ref{tab:ablation2e5}), are provided in Appendix~\ref{app:analyses}. \method{} is insensitive to the heal length $k$ (Table~\ref{tab:heal}), so one setting can transfer to every other model and split without retuning.

\begin{figure}[t]
\centering
\begin{tikzpicture}
\begin{axis}[width=\columnwidth, height=3.0cm, ymode=log, ymin=0.0018, ymax=0.15, xmin=0.7, xmax=6.3,
  ylabel={$\Delta$oracle after relearning $\downarrow$}, ylabel style={font=\scriptsize}, label style={font=\small},
  xtick={1,2,3,4,5,6}, xticklabels={1 ep.,2 ep.,3 ep.,1 ep.,2 ep.,3 ep.}, tick label style={font=\scriptsize},
  ytick={0.003,0.01,0.03,0.1}, yticklabels={$0.003$,$0.01$,$0.03$,$0.1$}, grid=major, grid style={gray!15},
  xlabel={attack epochs at lr $10^{-5}$ (left) and $2\cdot10^{-5}$ (right)}, xlabel style={font=\scriptsize},
  legend style={font=\scriptsize, draw=none, fill=none, at={(0.5,1.02)}, anchor=south, legend columns=3,
    /tikz/every even column/.append style={column sep=4pt}},
  every axis plot/.append style={line width=0.8pt, mark size=1.5pt},
  sweepNPO/.style={oiorange, mark=square*}, sweepUNDIAL/.style={gray, mark=diamond*},
  sweepWGA/.style={oipurple, mark=triangle*}, sweepRMU/.style={oiverm, mark=pentagon*},
  sweepDPO/.style={oisky, mark=x}, sweepSHIFT/.style={oiblue, line width=1.8pt, mark=*, mark size=2pt}]
\draw[densely dashed, gray!60] (axis cs:3.5,0.0018) -- (axis cs:3.5,0.15);
\input{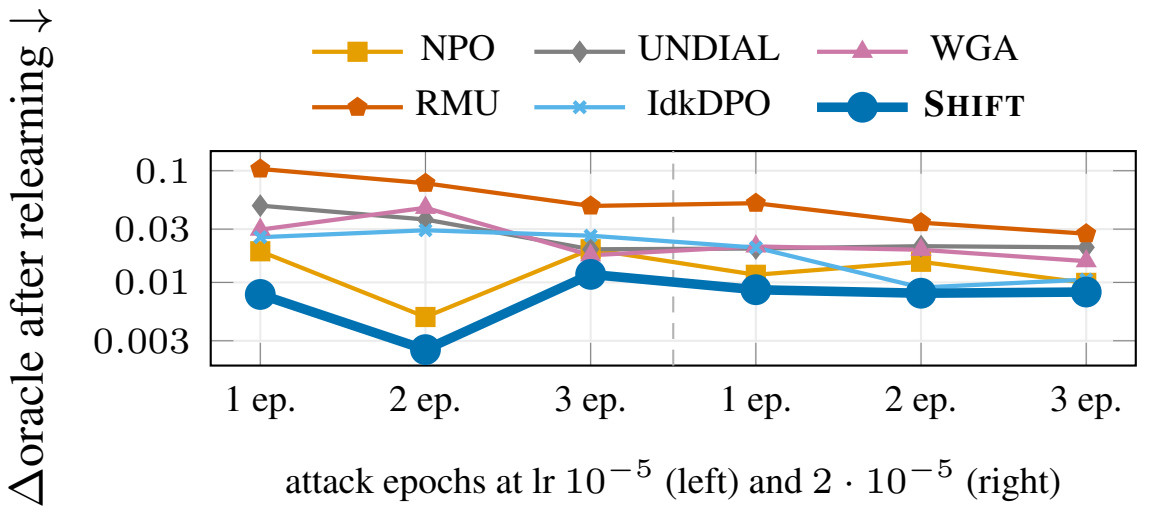}
\legend{NPO, UNDIAL, WGA, RMU, IdkDPO, \textbf{\method{}}}
\end{axis}
\end{tikzpicture}
\caption{Relearning attacks of increasing strength on Llama-3.2-1B \texttt{forget05}: distance to the equally attacked
oracle (mean of three attack seeds). \method{} is the closest at every strength (results for all baselines refer to Appendix Table~\ref{tab:attack-sweep}).}
\label{fig:sweep}
\end{figure}

\expbold{Robustness to relearning.} Suppressed knowledge can return after fine-tuning on related data
\citep{lynch2024eight,hu2025unlearning,deeb2024unlearning}. We attack the $2\cdot10^{-5}$ models, relearning half of the \texttt{forget05} authors and testing the other half (see details in Appendix~\ref{app:attack}). The results are reported in Figure~\ref{fig:sweep} and Appendix Table~\ref{tab:attack-sweep}. After the attack, \method{} is within $0.008$ of the equally attacked oracle, $2.4\times$ closer than any baseline, whereas aggressive forgetting bounces back (RMU's extraction strength rises from $0.029$ to $0.195$;
Appendix Table~\ref{tab:attack}), and \method{} stays the closest at all six attack strengths (Figure~\ref{fig:sweep}).

\section{Conclusion}

Likelihood-suppression unlearning has no target, so its outcome is set by the optimization budget. The shrink-and-heal lookahead estimates this target from the model itself (its linearized fixed point is the retrained oracle); \method{} distills it on forget tokens and iterates. Across three model sizes, it is on average the closest method to the oracle and has the highest average aggregate score, and it stays near the oracle where the baselines overshoot: unlearning should target an estimate of the oracle.

\clearpage
\section*{Limitations}

\method{} needs the pre-fine-tuning weights $\thb$; it applies to knowledge acquired in an identifiable
fine-tuning stage (as in TOFU and MUSE), not to knowledge from pre-training, where no earlier checkpoint without
it may exist. Our analysis assumes the linearized regime, a heal run to convergence and an exact teacher (no
clipping or top-token truncation); we measured where real models depart from it (Section~\ref{sec:probe}) but do not bound the
error.

\section*{Ethics Statement}

Unlearning is used to remove private or copyrighted content; methods that look successful on a single metric but
over-forget are detectable by membership inference, which itself leaks which data were removed
\citep{hayes2025inexact}. Calibrating forgetting to the retrained model is meant to reduce this leak. TOFU uses
fictitious authors; MUSE uses public news articles. No new personal data were collected.

\bibliography{references}

\clearpage
\appendix
\renewcommand{\dbltopfraction}{0.9}
\renewcommand{\topfraction}{0.9}
\renewcommand{\textfraction}{0.07}
\renewcommand{\dblfloatpagefraction}{0.8}
\renewcommand{\floatpagefraction}{0.8}

This appendix provides additional related work, theoretical proofs, implementation details, and experimental results supporting the main paper. Specifically, it includes:
\begin{itemize}[leftmargin=*,itemsep=1pt,topsep=2pt,noitemsep]
    \item Appendix~\ref{app:related} discusses extended related work: further discussion of unlearning objectives, oracle-based evaluation, distillation, robustness, and model arithmetic.
    \item Appendix~\ref{app:proofs} provides proofs: complete proofs of Propositions~\ref{prop:probe} and~\ref{prop:distill}, Corollary~\ref{cor:iter}, and Lemma~\ref{lem:norm}, along with an analysis of finite healing.
    \item Appendix~\ref{app:stage0} measures the Lookahead: empirical analysis of the lookahead's effects on forget-set and unseen examples.
    \item Appendix~\ref{app:impl} presents the implementation details, including training configurations, hyperparameters, and the complete algorithm.
    \item Appendix~\ref{app:full} provides full TOFU results, including detailed memorization, utility, and aggregate scores across TOFU settings.
    \item Appendix~\ref{app:analyses} reports additional analyses results, including learning-rate sensitivity, heal-length sensitivity, and ablations at larger learning rate.
    \item Appendix~\ref{app:attack} discusses relearning Attack, including attack setup and evaluation of oracle alignment under varying relearning strengths.
\end{itemize}

\section{Extended Related Work}
\label{app:related}

\paragraph{Likelihood suppression.} Early unlearning methods use gradient ascent on the forget set~\citep{jang2023knowledge,yao2024large}, often combined with retain-set regularization~\citep{liu2022continual}. Subsequent approaches introduce bounded or reweighted objectives~\citep{zhang2024negative,fan2024simplicity,
wang2025rethinking,yang2025exploring}, refusal targets~\citep{maini2024tofu}, representation misdirection~\citep{li2024wmdp}, and constrained optimization~\citep{entesari2025constrained}. Recent work also explores adaptive stopping criteria based on the model's own predictions~\citep{digennaro2026fast}. Despite their different mechanisms, they do not explicitly target the behavior of the retain-only oracle, which provides a principled reference for unlearning~\citep{yoon2026position,hayes2025inexact}.

\paragraph{Evaluation and scope.} Recent position and evaluation work argues that unlearning claims should be judged
against retraining \citep{yoon2026position} and that forgetting must generalize narrowly while retention generalizes broadly
\citep{peleg2026forget}; robustness to relearning depends on how forget- and retain-critical weights are separated
\citep{chen2026distance}. Our evaluation follows the first view: we measure the distance to the retrained model.
Measuring the gap to the retrained model is standard beyond LLMs: forgetting has been defined as indistinguishability
from a model trained without the data \citep{golatkar2020eternal}, image-classification unlearning reports the average gap
to Retrain across accuracy and membership-inference metrics \citep{fan2024salun}, and SCRUB adds a rewinding step for
privacy so that the forget set is not forgotten beyond what retraining would give \citep{kurmanji2023towards}. For LLMs,
TOFU's forget quality tests whether the forget-set truth ratios are distributed as the retain model's
\citep{maini2024tofu}, and MUSE's PrivLeak compares the membership-inference AUC with the retrained model's, flagging both
under- and over-unlearning \citep{shi2025muse}. Our $\Delta$ applies this gap to each forget statistic, PL carries
MUSE's measure over to TOFU, and Table~\ref{tab:fidelity} reports TOFU's forget quality.

\paragraph{Distillation and target distributions.} Teacher--student unlearning uses an incompetent or
adjusted teacher \citep{chundawat2023can,kurmanji2023towards}; UNDIAL lowers the gold logit by a fixed amount
\citep{dong2025undial}; ``Who's Harry Potter'' builds the teacher
from a model \emph{reinforced} on the forget data and subtracts its excess \citep{eldan2023whos}. \method{} instead
moves \emph{away} from the fine-tuning that produced the knowledge, and calibrates the move by the retain set.
Logit-difference methods combine the model with an assistant trained to remember the forget set
\citep{ji2024reversing,huang2024offset}; emulated fine-tuning adds or removes the effect of fine-tuning at the logit
level \citep{mitchell2024emulator}. DUET distills from an in-context prompted teacher \citep{zhong2026duet}, and bootstrapping
methods suppress the model's own high-confidence generations to avoid the squeezing of probability mass into rephrasings
\citep{li2026llm}. Simulated oracle matching predicts the oracle's outputs from data attribution
\citep{georgiev2024attribute}; \method{} predicts them from a cheap retain re-fit.

\paragraph{Robustness of unlearning.} Suppressed knowledge can return after fine-tuning on related data
\citep{lynch2024eight,hu2025unlearning,deeb2024unlearning}, especially when the relearning data share their format with
the forget set \citep{yoon2026rethinking}, or after low-bit quantization \citep{zhang2025catastrophic}. Recent defenses
target the answer margins \citep{yin2026crossing}, minor representation components \citep{xiao2026robust}, smoothness in
parameter and representation space \citep{yan2026dual}, or quantization explicitly \citep{sadhu2026forgetting}, and
distilling toward a target without the knowledge has been argued to remove it more durably \citep{lee2025distillation}.
\method{} is a distillation method of this kind, with a target estimated from the retain set.

\paragraph{Checkpoint arithmetic.} The closest related approach is MSA \citep{rezaei2026revisiting}, which also uses a checkpoint from before the forget set was seen, but in the opposite direction: it fine-tunes that checkpoint \emph{on} $\Df$ and subtracts the resulting vector, scaled by a tuned coefficient, from the final model. \method{} never trains on $\Df$. It shrinks toward the checkpoint and re-fits $\Dr$, so the forget direction is what $\Dr$ cannot restore, and its magnitude is set by the oracle (Proposition~\ref{prop:probe}) rather than by that coefficient. The estimate is then read on forget tokens and distilled, instead of being subtracted in weight space.

\paragraph{Resetting weights.} Shrink-and-perturb \citep{ash2020warm}, rewinding to an earlier checkpoint and
fine-tuning on retained data \citep{mu2024rewind}, Fisher-based scrubbing \citep{golatkar2020eternal} and
noising followed by distillation \citep{lee2025distillation} all \emph{commit} a reset. SAGE projects an
unlearning update away from directions that damage retain activations \citep{zhang2026sage}, close in spirit to how our
heal restores retain-supported directions. Our ablation that keeps the lookahead as the model (Table~\ref{tab:ablation}) belongs to
this family; \method{} reads the reset instead of keeping it, and derives a calibrated target from it. Task
arithmetic negates a task vector to forget a task \citep{ilharco2023editing}; Proposition~\ref{prop:probe} explains why
shrinking and re-fitting the retain set isolates the forget-specific part of that vector without training on the
forget set.

\section{Proofs}
\label{app:proofs}

\paragraph{Setup.} Let $f_X(\theta)\in\R^{m}$ stack the logits of a set $X$ of training positions. In the linearized model,
\begin{equation*}
f_X(\theta)=f_X(\thb)+J_X\,(\theta-\thb).
\end{equation*}
Training to interpolation fits targets $y_X$ (for the cross-entropy loss, logits that realize the fitted distributions; the argument only needs that the fitted values are reached). We use one standard fact about gradient descent~\citep{gunasekar2018characterizing}:
\begin{tcolorbox}[colback=blue!2,colframe=blue!20,left=1mm,right=1mm,top=1mm,bottom=1mm, before skip=8pt, after skip=8pt]
\textbf{Fact.}
Gradient descent on any loss of $f_X$, started from $\theta'$, moves only within the row space
$\mathrm{row}(J_X)$. If it fits $y_X$, it therefore converges to the closest fitting point:
\begin{equation}
\theta'+J_X^{+}\big(y_X-f_X(\theta')\big).
\label{eq:min-norm}
\end{equation}
\end{tcolorbox}

\noindent For the retain set we write $u_r=y_r-f_r(\thb)$ for the residual of the base model,
$\proj=J_r^{+}J_r$ for the projection onto $\mathrm{row}(J_r)$, and $\tau=\tho-\thb$ for what fine-tuning added.

\begin{proof}[\normalfont\bfseries Proof of Proposition~\ref{prop:probe}]
\emph{Step 1: the oracle.} The target model fits $\Dr$, so $J_r\tau=u_r$. The oracle is gradient descent on $\Dr$ from
$\thb$, so by (F)
\begin{equation*}
\thr=\thb+J_r^{+}u_r=\thb+J_r^{+}J_r\tau=\thb+\proj\tau,
\end{equation*}
which is Eq.~(\ref{eq:oracle}).

\emph{Step 2: the shrink.} The shrunk model $\theta'=\thb+(1-\rho)\tau$ leaves a fraction $\rho$ of the retain set to be
re-learned:
\begin{equation*}
y_r-f_r(\theta')=u_r-(1-\rho)J_r\tau=\rho\,u_r .
\end{equation*}

\emph{Step 3: the heal.} By (F), healing from $\theta'$ adds $J_r^{+}(\rho\,u_r)=\rho\,\proj\tau$. Hence
\begin{align*}
\Pi(\tho)&=\thb+(1-\rho)\tau+\rho\,\proj\tau,\\
\Pi(\tho)-\tho&=-\rho\,(I-\proj)\tau=\rho\,(\thr-\tho),
\end{align*}
where the last equality uses $\thr-\tho=\proj\tau-\tau$ from Step 1. This is Eq.~(\ref{eq:probe}).

\emph{Step 4: the fixed point.} Every lookahead fits $\Dr$, because the heal runs to convergence, so it suffices to
consider $\theta=\thb+\tau'$ with $J_r\tau'=u_r$. Steps 2--3 then apply with $\tau'$ in place of $\tau$:
\begin{equation*}
\Pi(\theta)-\theta=-\rho\,(I-\proj)\tau' .
\end{equation*}
A fixed point has $(I-\proj)\tau'=0$, i.e., $\tau'\in\mathrm{row}(J_r)$ with $J_r\tau'=u_r$. The only such vector is
$\tau'=J_r^{+}u_r=\proj\tau$, so the fixed point is $\thr$, and it is unique. This completes the proof.
\end{proof}

\begin{proof}[\normalfont\bfseries Proof of Proposition~\ref{prop:distill}]
Let $J_{\mathcal D}$ stack the Jacobians of all training positions of $\Df\cup\Dr$, write $\tau_n=\theta_n-\thb$, and
let
\begin{equation*}
\theta^\star=\theta_n+\tfrac{s}{\rho}\big(\Pi(\theta_n)-\theta_n\big)=\theta_n-s\,(I-\proj)\tau_n ,
\end{equation*}
where the second form is Proposition~\ref{prop:probe}. We show that distillation from $\theta_n$ lands exactly on
$\theta^\star$.

\emph{Step 1: the targets are the outputs of $\theta^\star$.} On $\Df$, linearity (and Lemma~\ref{lem:norm} for
log-probabilities) gives
\begin{equation*}
f_f(\theta_n)+\tfrac{s}{\rho}\big(f_f(\Pi(\theta_n))-f_f(\theta_n)\big)=f_f(\theta^\star).
\end{equation*}
On $\Dr$, the anchor's target is $f_r(\tho)=y_r$, and $f_r(\theta^\star)=f_r(\theta_n)=y_r$ because
$J_r(I-\proj)=0$.

\emph{Step 2: the step lies in $\mathrm{row}(J_{\mathcal D})$.} The target model is the minimum-norm fit of
$\mathcal D$, so $\tau\in\mathrm{row}(J_{\mathcal D})$. Each iterate is
$\tau_{n+1}=(1-s)\,\tau_n+s\,\proj\tau_n$ with $\proj\tau_n\in\mathrm{row}(J_r)\subseteq\mathrm{row}(J_{\mathcal D})$,
so by induction from $\tau_0=\tau$ every $\tau_n$ lies in $\mathrm{row}(J_{\mathcal D})$, and hence so does
$\theta^\star-\theta_n=-s\,(I-\proj)\tau_n$.

\emph{Step 3: conclusion.} By (F), gradient descent on the logit-matching loss from $\theta_n$ converges to
\begin{align*}
&\theta_n+J_{\mathcal D}^{+}\big(f_{\mathcal D}(\theta^\star)-f_{\mathcal D}(\theta_n)\big)\\
&\qquad=\theta_n+J_{\mathcal D}^{+}J_{\mathcal D}\,(\theta^\star-\theta_n)=\theta^\star ,
\end{align*}
since $J_{\mathcal D}^{+}J_{\mathcal D}$ is the identity on $\mathrm{row}(J_{\mathcal D})$. At $s=1$,
Corollary~\ref{cor:iter} gives $\theta^\star=\thr$. The KL form of the loss is minimized by any logits that equal the
targets up to a constant per position, and those leave the predictive distributions unchanged.
\end{proof}

\begin{proof}[\normalfont\bfseries Proof of Corollary~\ref{cor:iter}]
Write $\tau_n=\theta_n-\thb$. If $\theta_n$ fits $\Dr$, Proposition~\ref{prop:probe} gives
\begin{align*}
\theta_{n+1}&=\theta_n+\tfrac{s}{\rho}\big(\Pi(\theta_n)-\theta_n\big)\\
&=\theta_n-s\,(I-\proj)\tau_n .
\end{align*}
The step lies in the null space of $J_r$, so $J_r\tau_{n+1}=J_r\tau_n=u_r$ and $\theta_{n+1}$ still fits $\Dr$; by
induction from $\theta_0=\tho$, every iterate does. Projecting the step onto $\mathrm{row}(J_r)$ and its complement,
\begin{align*}
\proj\tau_{n+1}&=\proj\tau_n,\\
(I-\proj)\tau_{n+1}&=(1-s)\,(I-\proj)\tau_n .
\end{align*}
Unrolling from $\tau_0=\tau$,
\begin{equation*}
\proj\tau_n=\proj\tau,\qquad (I-\proj)\tau_n=(1-s)^{n}(I-\proj)\tau .
\end{equation*}
For $s\in(0,2)$ we have $|1-s|<1$, so $\tau_n\to\proj\tau$, i.e., $\theta_n\to\thr$; for $s=1$ the forget-specific
part vanishes after one step.
\end{proof}

\begin{lemma}[Normalization invariance]\label{lem:norm}
For logits $z,z'$ with log-probabilities $\ell=z-\mathrm{lse}(z)$ and $\ell'=z'-\mathrm{lse}(z')$, and any $\beta$,
\[
\mathrm{softmax}\big(\ell+\beta(\ell'-\ell)\big)=\mathrm{softmax}\big(z+\beta(z'-z)\big).
\]
\end{lemma}

\begin{proof}[\normalfont\bfseries Proof of Lemma~\ref{lem:norm}]
Substituting the definitions,
\begin{align*}
\ell+\beta(\ell'-\ell)&=z+\beta(z'-z)\\
&\quad-\big[(1-\beta)\,\mathrm{lse}(z)+\beta\,\mathrm{lse}(z')\big].
\end{align*}
The bracket is the same across tokens, and softmax is invariant to adding a constant to all logits.
\end{proof}

\paragraph{A finite heal.} With $k$ heal steps of size $\eta$ on the squared loss, the healed component along an
eigenvector $v_i$ of $J_r^\top J_r$ with eigenvalue $\lambda_i$ is a fraction $1-(1-\eta\lambda_i)^k$ of its shrunk
part. The lookahead step therefore becomes
\begin{align*}
\Pi(\tho)-\tho&=-\rho\,(I-\proj)\tau\\
&\quad-\rho\sum_i(1-\eta\lambda_i)^k\,v_iv_i^\top\proj\tau .
\end{align*}
The first term is the step of Proposition~\ref{prop:probe}. The second is the retain knowledge that the heal did not
restore; it lies in retain-supported directions, which is what the retain anchor of Eq.~(\ref{eq:loss}) counteracts
(Section~\ref{sec:iterate}).

\section{Measuring The Lookahead}
\label{app:stage0}

Before building the method, we measured the lookahead at $\tho$ (TOFU \texttt{forget10}, Llama-3.2-1B, $\rho=\tfrac12$). For each answer position, $d_t=-(\ell_\Pi-\ell_0)(y_t)/\rho$ is the extrapolated ($s=1$) drop of the gold token's log-probability: positive when the lookahead lowers the gold token, negative when it raises it. On never-seen holdout authors (positions 200--300 of \texttt{holdout10}) we fit a band of $d$ for each confidence decile of $\ell_0(y_t)$, and call a position \emph{active} when its $d$ falls outside the band. $R$ is the gated forget-set drop at $s=1$ (forget minus null).

\begin{table}[H]
\centering
\footnotesize
\setlength{\tabcolsep}{3pt}
\begin{adjustbox}{max width=\columnwidth}
\begin{tabular}{lrrrr}
\toprule
Heal steps $k$ & $\bar d$ holdout & $\bar d$ forget & $R$ & Active F\,/\,null\,/\,test \\
\midrule
16 & $-0.876$ & $+1.414$ & 1.58 & 0.26\,/\,0.05\,/\,0.056 \\
32 & $-0.955$ & $+1.489$ & 1.93 & 0.31\,/\,0.05\,/\,0.058 \\
64 & $-0.933$ & $+1.542$ & 1.84 & 0.30\,/\,0.05\,/\,0.060 \\
\bottomrule
\end{tabular}
\end{adjustbox}
\caption{The lookahead at $\tho$: mean extrapolated drop $\bar d$ (nats/token) on never-seen and forget authors, gated
forget-specific drop $R$, and the fraction of active positions on the forget set, the band's own authors (null) and a
second set of never-seen authors (test).}
\label{tab:stage0}
\end{table}

\noindent The lookahead lowers the gold tokens of the forget set ($\bar d>1.4$ nats) and raises them on
never-seen authors (by $0.88$--$0.96$ nats). The band transfers to the second set of never-seen authors (5.6--6.0\% active, as designed), where
the gated drop is 0.04--0.08 nats/token. For reference, $\tho$'s log-probability of the gold forget tokens exceeds the
retrained model's by $2.45$ nats/token (computed post hoc; the method never uses the retrained model). The heal does not
remove the drift on holdout authors between 16 and 64 steps, so a longer heal would not remove it either; this is why
\method{} confines the teacher to forget tokens rather than trying to make the lookahead exact.

\section{Implementation Details}
\label{app:impl}

\subsection{Implementation Setup}

\paragraph{Environment.} Llama-3.2-1B runs on one NVIDIA GeForce RTX 5060 Ti (16\,GB) or A100 (40\,GB), Llama-3.2-3B on
one A100, and Llama-3.1-8B and the MUSE model on four A100 GPUs with DeepSpeed ZeRO-3 (optimizer states on the CPU).
All code builds on OpenUnlearning \citep{dorna2025openunlearning} with PyTorch 2.4.1, Transformers 4.51.3 and DeepSpeed
0.15.4.

\paragraph{Training configuration.} Every method follows the OpenUnlearning configuration. On TOFU we use AdamW (8-bit
paged optimizer states), an effective batch size of 32, a learning rate of $10^{-5}$ with one epoch of linear warm-up and
linear decay, weight decay 0.01, 10 epochs and bf16 weights; with Llama-3.1-8B every method keeps the input embeddings
frozen. On MUSE we use an effective batch size of 64, a learning rate of $10^{-5}$ with a constant schedule, 10 epochs
and no weight decay. Each method is run once (seed 0) and its final model is evaluated, with no checkpoint selection.

\subsection{Hyperparameter Settings}

Every baseline keeps the OpenUnlearning defaults below ($\gamma$: weight of the forget loss, $\alpha$: weight of the
retain loss, NLL on the retain set unless noted); we do not tune them per setting.
\begin{itemize}\setlength{\itemsep}{1pt}\setlength{\parskip}{0pt}
\item \textbf{GradAscent.} No method-specific hyperparameter.
\item \textbf{GradDiff.} $\gamma=\alpha=1$.
\item \textbf{NPO.} $\beta=0.1$, $\gamma=\alpha=1$.
\item \textbf{SimNPO.} $\beta=4.5$, margin $\delta=0$, $\gamma=0.125$, $\alpha=1$.
\item \textbf{IdkDPO.} $\beta=0.1$ with refusal answers as the preferred responses, $\gamma=\alpha=1$.
\item \textbf{RMU.} Steering coefficient 2 at layer 7, all parameters trained, retain loss on hidden representations,
$\gamma=\alpha=1$.
\item \textbf{UNDIAL.} $\beta=10$, $\gamma=1$, $\alpha=0$.
\item \textbf{WGA.} $\beta=1$, $\gamma=\alpha=1$.
\item \textbf{SatImp.} $\beta_1=5$, $\beta_2=1$, $\gamma=0.1$, $\alpha=1$.
\item \textbf{CE-U.} The first answer token is not trained.
\item \textbf{PDU.} Primal-dual updates with dual step size 1 and 5 warm-up epochs, retain tolerance $\varepsilon=0$,
$\gamma=\alpha=1$.
\item \textbf{\method{}.} Lookahead: $\rho=\tfrac12$ and $k=16$ heal steps on retain answers (32 QA pairs per step,
learning rate $10^{-5}$). Teacher: the top-256 tokens of the model and the lookahead at each position (top-128 on MUSE),
clip $\kappa=10$, step sizes $s=(\tfrac12,1,1,1)$. Distillation: $\gamma=\alpha=1$, a retain anchor on $\tho$'s top-64
next-token distributions, learning rate $10^{-5}$ with a short warm-up in each round, four rounds of four epochs.
\end{itemize}

\begin{algorithm}[H]
\caption{\method{}}\label{alg:shift}
\small
\algrenewcommand{\algorithmiccomment}[1]{\hfill{\color{gray!75!black}\(\triangleright\)~#1}}
\begin{algorithmic}[1]
\Require target $\tho$, base $\thb$, $\Df$, $\Dr$; $\rho$, $k$, $\kappa$, $s_0,\dots,s_{N-1}$
\State cache $p_{\tho}$ on $\Dr$ \Comment{retain anchor}
\For{$n=0,\dots,N-1$}
  \State $\Pi\gets\mathrm{Heal}_k\big(\theta_n-\rho\,(\theta_n-\thb)\big)$ \Comment{Eq.~(\ref{eq:probe-def})}
  \ForAll{answer positions $t$ of $\Df$}
    \State $\delta_t\gets\ell_{\Pi,t}-\ell_{n,t}$ \Comment{on top tokens $V_t$}
    \State $p_{T,t}\gets\mathrm{norm}_{V_t}\big(\ell_{n,t}+\mathrm{clip}_{\kappa}(\tfrac{s_n}{\rho}\delta_t)\big)$ \Comment{Eq.~(\ref{eq:teacher})}
  \EndFor
  \State cache $p_T$ on $\Df$; free $\Pi$
  \State $\theta_{n+1}\gets$ distill from $\theta_n$ on $\mathcal{L}$ \Comment{Eq.~(\ref{eq:loss})}
\EndFor
\State \Return $\theta_N$
\end{algorithmic}
\end{algorithm}

\section{Full TOFU Results}
\label{app:full}

Table~\ref{tab:overview} complements Table~\ref{tab:agg} with the two components of Agg. \method{}'s memorization and
utility both stay close to the oracle's, whereas on average every baseline gives up one of the two: most leave the forget set
memorized (low Mem.), and GradAscent and CE-U collapse (Util.\ near $0$).

\begin{table*}[tp]
\centering
\footnotesize
\setlength{\tabcolsep}{4pt}
\begin{adjustbox}{max width=\textwidth}\begin{tabular}{lrrrrrrrrr}
\toprule
 & \multicolumn{3}{c}{Llama-3.2-1B} & \multicolumn{3}{c}{Llama-3.2-3B} & \multicolumn{3}{c}{Llama-3.1-8B} \\
\cmidrule(lr){2-4} \cmidrule(lr){5-7} \cmidrule(lr){8-10}
Method & Agg.$\uparrow$ & Mem.$\uparrow$ & Util.$\uparrow$ & Agg.$\uparrow$ & Mem.$\uparrow$ & Util.$\uparrow$ & Agg.$\uparrow$ & Mem.$\uparrow$ & Util.$\uparrow$ \\
\midrule
\multicolumn{10}{c}{\textsc{Forget 1\%}} \\
\midrule
\textit{Full} & 0.118 & 0.064 & 0.717 & 0.024 & 0.012 & 0.763 & 0.005 & 0.002 & 0.738 \\
\textit{Retain} & 0.613 & 0.543 & 0.703 & 0.632 & 0.541 & 0.761 & 0.619 & 0.533 & 0.738 \\
GradAscent & 0.458 & 0.335 & 0.723 & 0.409 & 0.281 & 0.750 & 0.617 & 0.542 & 0.716 \\
GradDiff & 0.440 & 0.317 & 0.718 & 0.378 & 0.251 & 0.760 & 0.565 & 0.475 & 0.698 \\
NPO & 0.458 & 0.336 & 0.717 & 0.419 & 0.290 & 0.750 & 0.599 & 0.504 & 0.737 \\
SimNPO & 0.218 & 0.129 & 0.725 & 0.151 & 0.084 & 0.746 & 0.250 & 0.150 & 0.762 \\
IdkDPO & 0.294 & 0.185 & 0.712 & 0.196 & 0.112 & 0.762 & 0.478 & 0.365 & 0.693 \\
RMU & 0.509 & 0.414 & 0.660 & 0.212 & 0.123 & 0.768 & 0.533 & 0.419 & 0.734 \\
UNDIAL & 0.456 & 0.335 & 0.718 & 0.446 & 0.316 & 0.759 & \underline{0.630} & 0.538 & 0.759 \\
WGA & 0.434 & 0.310 & 0.723 & 0.405 & 0.277 & 0.753 & 0.552 & 0.436 & 0.751 \\
SatImp & 0.139 & 0.077 & 0.722 & 0.036 & 0.019 & 0.756 & 0.106 & 0.057 & 0.762 \\
CE-U & 0.557 & 0.466 & 0.692 & 0.538 & 0.417 & 0.761 & 0.282 & 0.796 & 0.171 \\
PDU & \underline{0.573} & 0.488 & 0.692 & \underline{0.559} & 0.459 & 0.714 & 0.018 & 0.810 & 0.009 \\
\cmidrule(lr){1-10}
\textbf{SHIFT (ours)} & \textbf{0.619} & 0.542 & 0.720 & \textbf{0.635} & 0.546 & 0.758 & \textbf{0.639} & 0.565 & 0.736 \\
\midrule
\multicolumn{10}{c}{\textsc{Forget 5\%}} \\
\midrule
\textit{Full} & 0.154 & 0.086 & 0.707 & 0.059 & 0.030 & 0.748 & 0.016 & 0.008 & 0.723 \\
\textit{Retain} & 0.641 & 0.577 & 0.722 & 0.648 & 0.569 & 0.751 & 0.641 & 0.567 & 0.736 \\
GradAscent & 0.214 & 0.635 & 0.129 & \underline{0.567} & 0.501 & 0.653 & 0.000 & 0.983 & 0.000 \\
GradDiff & 0.527 & 0.475 & 0.591 & 0.487 & 0.399 & 0.624 & 0.111 & 0.988 & 0.059 \\
NPO & 0.548 & 0.507 & 0.597 & 0.561 & 0.477 & 0.681 & \textbf{0.666} & 0.609 & 0.736 \\
SimNPO & 0.229 & 0.136 & 0.709 & 0.165 & 0.093 & 0.741 & 0.366 & 0.244 & 0.734 \\
IdkDPO & 0.442 & 0.325 & 0.689 & 0.397 & 0.272 & 0.734 & 0.549 & 0.453 & 0.696 \\
RMU & \underline{0.575} & 0.524 & 0.638 & 0.539 & 0.421 & 0.748 & 0.612 & 0.729 & 0.527 \\
UNDIAL & 0.480 & 0.369 & 0.684 & 0.532 & 0.411 & 0.751 & 0.649 & 0.573 & 0.748 \\
WGA & 0.488 & 0.370 & 0.715 & 0.453 & 0.324 & 0.748 & 0.589 & 0.733 & 0.493 \\
SatImp & 0.175 & 0.100 & 0.710 & 0.080 & 0.042 & 0.740 & 0.143 & 0.080 & 0.713 \\
CE-U & 0.000 & 0.785 & 0.000 & 0.000 & 0.789 & 0.000 & 0.000 & 0.788 & 0.000 \\
PDU & 0.335 & 0.640 & 0.227 & 0.045 & 0.737 & 0.023 & 0.631 & 0.602 & 0.664 \\
\cmidrule(lr){1-10}
\textbf{SHIFT (ours)} & \textbf{0.623} & 0.558 & 0.705 & \textbf{0.639} & 0.556 & 0.752 & \underline{0.650} & 0.612 & 0.693 \\
\midrule
\multicolumn{10}{c}{\textsc{Forget 10\%}} \\
\midrule
\textit{Full} & 0.159 & 0.089 & 0.706 & 0.063 & 0.033 & 0.752 & 0.011 & 0.006 & 0.727 \\
\textit{Retain} & 0.643 & 0.584 & 0.715 & 0.650 & 0.574 & 0.749 & 0.649 & 0.573 & 0.748 \\
GradAscent & 0.000 & 0.992 & 0.000 & 0.000 & 0.968 & 0.000 & 0.000 & 0.982 & 0.000 \\
GradDiff & 0.521 & 0.494 & 0.551 & 0.482 & 0.474 & 0.490 & 0.423 & 0.988 & 0.269 \\
NPO & \underline{0.568} & 0.546 & 0.592 & 0.584 & 0.502 & 0.697 & \underline{0.650} & 0.597 & 0.714 \\
SimNPO & 0.235 & 0.141 & 0.709 & 0.181 & 0.103 & 0.748 & 0.383 & 0.261 & 0.715 \\
IdkDPO & 0.488 & 0.375 & 0.699 & 0.431 & 0.303 & 0.743 & 0.531 & 0.421 & 0.720 \\
RMU & 0.221 & 0.710 & 0.131 & \textbf{0.646} & 0.644 & 0.648 & 0.447 & 0.773 & 0.314 \\
UNDIAL & 0.484 & 0.374 & 0.686 & 0.546 & 0.426 & 0.757 & \textbf{0.653} & 0.583 & 0.742 \\
WGA & 0.506 & 0.392 & 0.711 & 0.485 & 0.360 & 0.743 & 0.436 & 0.771 & 0.304 \\
SatImp & 0.182 & 0.104 & 0.709 & 0.082 & 0.044 & 0.743 & 0.207 & 0.121 & 0.699 \\
CE-U & 0.000 & 0.785 & 0.000 & 0.000 & 0.794 & 0.000 & 0.000 & 0.990 & 0.000 \\
PDU & 0.499 & 0.410 & 0.638 & 0.520 & 0.434 & 0.647 & 0.310 & 0.802 & 0.192 \\
\cmidrule(lr){1-10}
\textbf{SHIFT (ours)} & \textbf{0.624} & 0.556 & 0.710 & \underline{0.634} & 0.550 & 0.748 & 0.631 & 0.581 & 0.691 \\
\bottomrule
\end{tabular}
\end{adjustbox}
\caption{Agg., Mem.\ and Util.\ on TOFU, all methods at learning rate $10^{-5}$ (one seed).
\textbf{Bold} / \underline{underlined}: best / second-best Agg.\ for each model and split; we do not mark Mem.\ or Util.\
alone, since collapsing maximizes Mem.\ (GradAscent and CE-U reach Util.\ $\approx0$).}
\label{tab:overview}
\end{table*}

\section{Additional Analyses}
\label{app:analyses}

\begin{table*}[t]
\centering
\footnotesize
\begin{minipage}[t]{0.61\textwidth}
\centering
\setlength{\tabcolsep}{3.5pt}
\begin{adjustbox}{max width=\linewidth}\begin{tabular}{lccccc}
\toprule
Method & $5\cdot10^{-6}$ & $10^{-5}$ & $2\cdot10^{-5}$ & $3\cdot10^{-5}$ & $5\cdot10^{-5}$ \\
\midrule
GradDiff & 0.353 / $-98$ & 0.527 / $-54$ & 0.539 / $+49$ & 0.293 / $+57$ & 0.094 / $+57$ \\
NPO & 0.335 / $-100$ & 0.548 / $-70$ & 0.621 / $+8$ & 0.638 / $+37$ & 0.631 / $+52$ \\
SimNPO & 0.179 / $-100$ & 0.229 / $-100$ & 0.344 / $-99$ & 0.420 / $-99$ & 0.486 / $-95$ \\
RMU & 0.446 / $-98$ & 0.575 / $-41$ & 0.182 / $+50$ & 0.099 / $+50$ & 0.085 / $+46$ \\
WGA & 0.253 / $-100$ & 0.488 / $-94$ & 0.634 / $-28$ & 0.681 / $+36$ & 0.692 / $+54$ \\
PDU & 0.451 / $-96$ & 0.335 / $-48$ & 0.016 / $+14$ & 0.121 / $+24$ & 0.503 / $+40$ \\
\midrule
\textbf{SHIFT (ours)} & 0.427 / $-97$ & 0.603 / $-74$ & 0.633 / $-23$ & 0.628 / $+6$ & 0.596 / $+6$ \\
\bottomrule
\end{tabular}
\end{adjustbox}
\caption{TOFU \texttt{forget05}, Llama-3.2-1B: Agg.\ / PL as the learning rate varies (seed 0; $10^{-5}$ is the
OpenUnlearning default run of each baseline; \method{} uses $k=32$ and two epochs a round at every rate). \method{}'s PL
stays within $[-23,+6]$ from $2\cdot10^{-5}$ up; at $5\cdot10^{-5}$ every baseline but SimNPO overshoots to $+40$--$+57$.}
\label{tab:lr}
\end{minipage}\hfill
\begin{minipage}[t]{0.35\textwidth}
\centering
\setlength{\tabcolsep}{4pt}
\begin{adjustbox}{max width=\linewidth}\begin{tabular}{lrrrr}
\toprule
$k$ & Agg.$\uparrow$ & Mem.$\uparrow$ & Util.$\uparrow$ & $\Delta$oracle$\downarrow$ \\
\midrule
0 & 0.617 & 0.553 & 0.698 & 0.015 \\
8 & 0.616 & 0.546 & 0.706 & 0.020 \\
\textbf{16} & 0.623 & 0.558 & 0.705 & 0.016 \\
32 & 0.619 & 0.551 & 0.706 & 0.016 \\
\bottomrule
\end{tabular}
\end{adjustbox}
\caption{TOFU \texttt{forget05}, Llama-3.2-1B, learning rate $10^{-5}$, four rounds of four epochs (one seed): \method{}
with $k$ heal steps per round. \textbf{Bold}: the value of the main results. $\Delta$oracle: as in Table~\ref{tab:fidelity}.}
\label{tab:heal}
\end{minipage}
\end{table*}

\begin{table*}[t]
\centering
\footnotesize
\setlength{\tabcolsep}{6pt}
\begin{adjustbox}{max width=\textwidth}\begin{tabular}{lcrrrr}
\toprule
Variant & $n$ & Agg.$\uparrow$ & Mem.$\uparrow$ & Util.$\uparrow$ & $\Delta$oracle$\downarrow$ \\
\midrule
\textbf{SHIFT} & 3 & 0.633$_{\pm .001}$ & 0.587$_{\pm .001}$ & 0.686$_{\pm .002}$ & \textbf{0.006} \\
\midrule
\quad no heal ($k{=}0$) & 3 & 0.632$_{\pm .002}$ & 0.584$_{\pm .001}$ & 0.689$_{\pm .004}$ & 0.011 \\
\quad no extrapolation ($s{=}\rho$) & 3 & 0.619$_{\pm .001}$ & 0.565$_{\pm .001}$ & 0.684$_{\pm .003}$ & \underline{0.007} \\
\quad no raised tokens & 3 & 0.582$_{\pm .000}$ & 0.506$_{\pm .001}$ & 0.686$_{\pm .001}$ & 0.041 \\
\midrule
Lookahead as the model, 1 round & 3 & 0.513$_{\pm .001}$ & 0.419$_{\pm .002}$ & 0.663$_{\pm .001}$ & 0.089 \\
Lookahead as the model, 4 rounds & 3 & 0.566$_{\pm .001}$ & 0.537$_{\pm .000}$ & 0.597$_{\pm .002}$ & 0.017 \\
\bottomrule
\end{tabular}
\end{adjustbox}
\caption{The ablations of Table~\ref{tab:ablation} at learning rate $2\cdot10^{-5}$ on TOFU \texttt{forget05},
Llama-3.2-1B ($n$: seeds). \textbf{Bold} / \underline{underlined}: closest / second closest to the oracle.}
\label{tab:ablation2e5}
\end{table*}

\paragraph{Learning rate.} Table~\ref{tab:lr} varies the learning rate of every method on Llama-3.2-1B
\texttt{forget05}. At the default $10^{-5}$ the baselines stop short of the oracle (privacy leakage far below 0); from
$2\cdot10^{-5}$ up most of them cross it, while \method{} stays near it.

\paragraph{Heal length.} Table~\ref{tab:heal} varies the number of heal steps $k$ on Llama-3.2-1B \texttt{forget05},
with everything else as in Table~\ref{tab:agg}. The results change little with $k$: Agg.\ stays between $0.616$ and
$0.623$ and the distance to the oracle between $0.015$ and $0.020$; $k=16$, the value of the main results, has the highest
Agg.

\paragraph{Ablations at the larger learning rate.} Table~\ref{tab:ablation2e5} repeats Table~\ref{tab:ablation} at
learning rate $2\cdot10^{-5}$ ($k=32$, two epochs a round, three seeds). With the larger update, the heal matters:
without it the distance to the oracle nearly doubles ($0.011$ vs.\ $0.006$), and \method{} is the closest variant.

\section{Relearning Attack}
\label{app:attack}

\begin{table*}[t]
\centering
\footnotesize
\begin{minipage}[t]{0.485\textwidth}
\centering
\setlength{\tabcolsep}{3pt}
\begin{adjustbox}{max width=\linewidth}\begin{tabular}{lrrrrr}
\toprule
 & \multicolumn{2}{c}{Before attack} & \multicolumn{2}{c}{After attack} & \\
\cmidrule(lr){2-3}\cmidrule(lr){4-5}
Method & ES & RL & ES & RL & $\Delta$oracle$\downarrow$ \\
\midrule
\textit{Full} & 0.684 & 0.831 & 0.316 & 0.537 & 0.214 \\
\textit{Retain} & 0.060 & 0.372 & 0.070 & 0.353 & -- \\
\midrule
GradAscent & 0.029 & 0.000 & 0.085 & 0.290 & 0.039 \\
GradDiff & 0.048 & 0.244 & 0.111 & 0.389 & 0.038 \\
NPO & 0.071 & 0.288 & 0.076 & 0.321 & \underline{0.019} \\
SimNPO & 0.301 & 0.591 & 0.253 & 0.489 & 0.160 \\
IdkDPO & 0.110 & 0.154 & 0.114 & 0.361 & 0.025 \\
RMU & 0.029 & 0.116 & 0.195 & 0.436 & 0.104 \\
UNDIAL & 0.047 & 0.310 & 0.103 & 0.418 & 0.049 \\
WGA & 0.062 & 0.305 & 0.112 & 0.371 & 0.030 \\
SatImp & 0.592 & 0.785 & 0.308 & 0.512 & 0.198 \\
CE-U & 0.029 & 0.001 & 0.078 & 0.271 & 0.045 \\
PDU & 0.029 & 0.073 & 0.123 & 0.340 & 0.033 \\
\midrule
\textbf{SHIFT (ours)} & 0.056 & 0.365 & 0.065 & 0.365 & \textbf{0.008} \\
\bottomrule
\end{tabular}
\end{adjustbox}
\caption{Relearning attack on TOFU \texttt{forget05} (Llama-3.2-1B, models unlearned at $2\cdot10^{-5}$): extraction
strength (ES) and ROUGE-L (RL) on the held-out half of the forget authors, before and after relearning on the other
half (mean of three attack seeds). $\Delta$oracle: mean absolute difference of ES and RL from the relearned
\textit{Retain} model; \textbf{bold} / \underline{underlined}: closest / second closest.}
\label{tab:attack}
\end{minipage}\hfill
\begin{minipage}[t]{0.485\textwidth}
\centering
\setlength{\tabcolsep}{3pt}
\begin{adjustbox}{max width=\linewidth}\begin{tabular}{lrrrrrr}
\toprule
 & \multicolumn{3}{c}{lr $10^{-5}$} & \multicolumn{3}{c}{lr $2\cdot10^{-5}$} \\
\cmidrule(lr){2-4}\cmidrule(lr){5-7}
Method & 1 ep. & 2 ep. & 3 ep. & 1 ep. & 2 ep. & 3 ep. \\
\midrule
GradAscent & 0.039 & 0.017 & \underline{0.016} & 0.017 & 0.022 & 0.014 \\
GradDiff & 0.038 & 0.048 & 0.023 & 0.029 & 0.021 & 0.013 \\
NPO & \underline{0.019} & \underline{0.005} & 0.020 & \underline{0.012} & 0.015 & \underline{0.010} \\
SimNPO & 0.160 & 0.092 & 0.065 & 0.055 & 0.038 & 0.032 \\
IdkDPO & 0.025 & 0.029 & 0.026 & 0.020 & \underline{0.009} & 0.011 \\
RMU & 0.104 & 0.077 & 0.048 & 0.051 & 0.034 & 0.027 \\
UNDIAL & 0.049 & 0.037 & 0.020 & 0.020 & 0.021 & 0.021 \\
WGA & 0.030 & 0.047 & 0.018 & 0.021 & 0.020 & 0.015 \\
SatImp & 0.198 & 0.111 & 0.077 & 0.061 & 0.036 & 0.024 \\
CE-U & 0.045 & 0.022 & 0.040 & 0.038 & 0.031 & 0.027 \\
PDU & 0.033 & 0.033 & 0.032 & 0.036 & 0.016 & 0.018 \\
\midrule
\textbf{SHIFT (ours)} & \textbf{0.008} & \textbf{0.002} & \textbf{0.012} & \textbf{0.009} & \textbf{0.008} & \textbf{0.008} \\
\bottomrule
\end{tabular}
\end{adjustbox}
\caption{Distance to the equally attacked oracle (mean absolute difference of ES and ROUGE-L on the held-out half)
across attack strengths, three attack seeds each. \textbf{Bold} / \underline{underlined}: closest / second
closest.}
\label{tab:attack-sweep}
\end{minipage}
\end{table*}

The attack fine-tunes the unlearned model on the first half of the \texttt{forget05} authors (TOFU has 20 consecutive
questions per author) with learning rate $10^{-5}$ for one epoch after 3 warmup steps (batch 8, fp32 master weights,
8-bit AdamW), and evaluates the second half, which it never sees. Results are averaged over three attack seeds,
which change the order of the relearning batches. Table~\ref{tab:attack-sweep} repeats the attack at learning rates
$10^{-5}$ and $2\cdot10^{-5}$ for one to three epochs (Figure~\ref{fig:sweep}). Stronger attacks also erode what the
full model itself knows about the held-out half, so every method moves toward the oracle, but \method{} stays the
closest at every strength.

\section{Use of AI Assistants}
\label{appx:llms}

In preparing this paper, we made limited use of AI Assistants such as ChatGPT only to help refine wording, correct grammar, and enhance clarity. All core research contributions, including conceptual development, theoretical analysis, methodology design and implementation, experimental evaluations, and paper organization and composition, were conducted entirely by the authors.

\end{document}